\documentclass[mnsc,nonblindrev]{informs3}

\OneAndAHalfSpacedXI 

\usepackage{natbib}
 \bibpunct[, ]{(}{)}{,}{a}{}{,}%
 \def\bibfont{\small}%
 \def\bibsep{\smallskipamount}%
 \def\bibhang{24pt}%

\usepackage{amsmath,amssymb, subcaption}
\usepackage{bm, bbm, booktabs}

\newcommand{\vx}{\mathbf{x}}

\newcommand{\vep}{\varepsilon}
\newcommand{\E}{\mathbb{E}}
\newcommand{\R}{\mathbb{R}}
\newcommand{\N}{\mathcal{N}}

\newcommand{\cO}{\mathcal{O}}
\newcommand{\tO}{\tilde{\mathcal{O}}}
\newcommand{\cE}{\mathcal{E}}
\newcommand{\cL}{\mathcal{L}}
\newcommand{\J}{\mathcal{J}}
\newcommand{\I}{\mathcal{I}}
\newcommand{\cN}{\mathfrak{N}}
\DeclareMathOperator{\tr}{tr}
\DeclareMathOperator{\var}{Var}

\newcommand{\hbeta}{\hat{\beta}}

\newcommand{\hdelta}{\hat{\delta}}
\newcommand{\tdelta}{\tilde{\delta}}

\newcommand{\betag}{\beta_{gold}^*}
\newcommand{\betap}{\beta_{proxy}^*}
\newcommand{\hbetag}{\hat{\beta}_{gold}}
\newcommand{\hbetap}{\hat{\beta}_{proxy}}
\newcommand{\Xg}{\mathbf{X}_{gold}}
\newcommand{\Xp}{\mathbf{X}_{proxy}}
\newcommand{\yg}{y_{gold}}
\newcommand{\yp}{y_{proxy}}
\newcommand{\Yg}{Y_{gold}}
\newcommand{\Yp}{Y_{proxy}}
\newcommand{\epg}{\varepsilon_{gold}}
\newcommand{\epp}{\varepsilon_{proxy}}
\newcommand{\Sigmag}{\Sigma_{gold}}
\newcommand{\Sigmap}{\Sigma_{proxy}}
\newcommand{\sigmag}{\sigma_{gold}}
\newcommand{\sigmap}{\sigma_{proxy}}
\newcommand{\ngold}{n_{gold}}
\newcommand{\nproxy}{n_{proxy}}

\newcommand{\blambda}{\bar{\lambda}}

\newcommand{\zero}{{\bf 0}}
\newcommand{\one}{{\bf 1}}

\newcommand{\bb}[1]{\left[#1\right]}
\newcommand{\bp}[1]{\left(#1\right)}
\newcommand{\bc}[1]{\left\{#1\right\}}
\newcommand{\bn}[1]{\left\|#1\right\|}

\TheoremsNumberedThrough     
\ECRepeatTheorems

\EquationsNumberedThrough    

\MANUSCRIPTNO{MS-0001-1922.65}

\begin{document}



\RUNTITLE{Predicting with Proxies}

\TITLE{Predicting with Proxies: \\Transfer Learning in High Dimension}

\ARTICLEAUTHORS{%
\AUTHOR{Hamsa Bastani}
\AFF{Wharton School, Operations Information and Decisions, \EMAIL{hamsab@wharton.upenn.edu}} 
} 

\ABSTRACT{%
Predictive analytics is increasingly used to guide decision-making in many applications. However, in practice, we often have limited data on the true predictive task of interest, and must instead rely on more abundant data on a closely-related \textit{proxy} predictive task. For example, e-commerce platforms use abundant customer click data (proxy) to make product recommendations rather than the relatively sparse customer purchase data (true outcome of interest); alternatively, hospitals often rely on medical risk scores trained on a different patient population (proxy) rather than their own patient population (true cohort of interest) to assign interventions. Yet, not accounting for the bias in the proxy can lead to sub-optimal decisions. Using real datasets, we find that this bias can often be captured by a sparse function of the features. Thus, we propose a novel two-step estimator that uses techniques from high-dimensional statistics to efficiently \textit{combine} a large amount of proxy data and a small amount of true data. We prove upper bounds on the error of our proposed estimator and lower bounds on several heuristics used by data scientists; in particular, our proposed estimator can achieve the same accuracy with exponentially less true data (in the number of features $d$). Our proof relies on a new LASSO tail inequality for approximately sparse vectors. Finally, we demonstrate the effectiveness of our approach on e-commerce and healthcare datasets; in both cases, we achieve significantly better predictive accuracy as well as managerial insights into the nature of the bias in the proxy data.
}%


\maketitle

%


\section{Introduction}

Decision-makers increasingly use machine learning and predictive analytics to inform consequential decisions. However, a pervasive problem that occurs in practice is the limited quantity of labeled data available in the desired setting. Building accurate predictive models requires significant quantities of labeled data, but large datasets may be costly or infeasible to obtain for the predictive task of interest. A common solution to this challenge is to rely on a \textit{proxy} --- a closely-related predictive task --- for which abundant data is already available. The decision-maker then builds and deploys a model predicting the proxy instead of the true task. To illustrate, consider the following two examples from revenue management and healthcare respectively:
\begin{example}[Recommendation Systems] \label{ex:expedia} A core business proposition for platforms (e.g., Expedia or Amazon) is to match customers with personalized product recommendations. The typical goal is to maximize the probability of a customer purchase by recommending products that a customer is most likely to purchase, based on past transaction data and customer purchase histories. Unfortunately, most platforms have sparse data on customer purchases (the true outcome they wish to predict) for a particular product, but significantly more data on customer \textit{clicks} (a proxy outcome). Clicks are a common proxy for purchases, since one may assume that customers will not click on a product without some intent to purchase. Consequently, platforms often recommend products with high predicted click-through rates rather than high predicted purchase rates.
\end{example}
\begin{example}[Medical Risk Scoring] \label{ex:diabetes} Many hospitals are interested in identifying patients who have high risk for some adverse event (e.g., diabetes, stroke) in order to target preventative interventions. This involves using past electronic medical records to train a patient \textit{risk score}, i.e., predict which patients are likely to get a positive diagnosis for the adverse event based on data from prior visits. However, small hospitals have limited data since their patient cohorts (true population of interest) are not sizable enough to have had a large number of adverse events. Instead, they adopt a published risk score trained on data from a larger hospital's patient cohort (proxy population). There are concerns that a predictive model trained at one hospital may not directly apply to a different hospital, since there are differences in physician behavior, patient populations, etc. Yet, one may assume that the large hospital's risk score is a good proxy for the small hospital's risk score, since the target of interest (the adverse event) is the same in both models.
\end{example}

There are numerous other examples of the use of proxies in practice. In Section \ref{ssec: more-proxies}, we overview the pervasiveness of proxies in healthcare and revenue management.

However, the use of proxies has clear drawbacks: the proxy and true predictive models may not be the same, and any bias between the two tasks will affect the predictive performance of the model. Consider Example \ref{ex:expedia} on recommendation systems. In Section \ref{ssec:expedia}, we use personalized hotel recommendation data from Expedia to demonstrate a systematic bias between clicks (proxy outcome) and purchases (true outcome). In particular, we find that the price of the recommendation negatively impacts purchases far more than clicks. Intuitively, a customer may not mind browsing expensive travel products, but is unlikely to make an expensive purchase. Thus, using predicted click-through rates alone (proxies) to make recommendations could result in overly expensive recommendations, thereby hurting purchase rates. Next, consider Example \ref{ex:diabetes} on medical risk scores. In Section \ref{ssec:diabetes}, we use electronic medical record data across several healthcare providers to demonstrate a systematic bias between a diabetes risk predictor trained on patient data from a large external hospital (proxy cohort) and a risk predictor trained on patient data from the small target hospital (true cohort). In particular, we find differences in physician diagnosing behavior (e.g., some physicians are more inclined than others to ask patients to fast in order to diagnose impaired fasting glucose) and how patient chart data is encoded in the medical record (e.g., obesity is recorded as a diagnosis more often in some hospitals despite patients having similar BMIs). As a result, features that are highly predictive in one hospital may not be predictive in another hospital, thereby hurting the performance of a borrowed (proxy) risk predictor at the target hospital.

We refer to data from the proxy and true predictive tasks as proxy and \textit{gold} data respectively. Analogously, we refer to estimators trained on proxy and gold data alone as the proxy and gold estimators respectively. Both estimators seek to predict outcomes on the true predictive task. From a statistical perspective, the gold estimator is unbiased but has high variance due to its limited sample size. On the other hand, the proxy estimator has low variance due to its large sample size, but may have a significant bias due to systematic differences between the true and proxy predictive tasks. Predictive accuracy is composed of both bias and variance. Thus, when we have a good proxy (the bias is not too large), the proxy estimator can be a much more accurate predictive model than the gold estimator, explaining the wide use of proxies in practice.


An immediate question is: can we \textit{combine} proxy and gold data to achieve a better bias-variance tradeoff and improve predictive accuracy? In many of these settings, we have access to (or could collect) information from \textit{both} predictive tasks, i.e., we typically have a large amount of proxy data \textit{and} a small amount of true data. For instance, platforms observe both clicks and purchases; the target hospital has access to both the published proxy estimator and basic summary statistics from an external hospital, as well as its own patient data. Thus, we have the opportunity to improve prediction by combining these data sources.
Conversations with professional data scientists indicate two popular heuristics: (i) model averaging over the gold and proxy estimators, and (ii) training a model on proxy and gold data simultaneously\footnote{One disadvantage of the weighted loss function is that it requires both proxy and gold data to be available together at the time of training. This may not be possible in settings such as healthcare, where data is sensitive.}, with a larger weight for gold observations. However, there is little understanding of whether and by how much these heuristics can improve predictive performance. Indeed, we prove lower bounds that both model averaging and weighted loss functions can only improve estimation error by at most a constant factor (beyond the naive proxy and gold estimators discussed earlier). Thus, neither approach can significantly improve estimation error.

Ideally, we would use the gold data to \textit{de-bias} the proxy estimator (which already has low variance); this would hopefully yield an estimator with lower bias while maintaining low variance. However, estimating the bias is challenging, as we have extremely limited gold data. In general, estimating the bias from gold data can be harder than directly estimating the true predictive model from gold data. Thus, we clearly need to impose additional structure to make progress.

Our key insight is that the bias between the true and proxy predictive tasks may often be well modeled by a \textit{sparse} function of the observed features. We argue that there is often some (a priori unknown) underlying mechanism that systematically affects a subset of the features, creating a bias between the true and proxy predictive tasks. When this is the case, we can successfully estimate the bias using high-dimensional techniques that exploit sparsity. To illustrate, we return to Examples \ref{ex:expedia} and \ref{ex:diabetes}. In the first example on hotel recommendations, we find on Expedia data that customers tend to click on more expensive products than they are willing to purchase. This creates a bias between the proxy and true predictive tasks that can be captured by the price feature alone. However, as we show in Fig. \ref{fig:expedia-coef-diff} in Section \ref{ssec:expedia}, the two predictive tasks appear remarkably similar otherwise. In particular, the difference of the proxy and gold estimators on Expedia data is very sparse (nearly all coefficients are negligible with the notable exception of the price coefficient). Similarly, in the second example on diabetes risk prediction, we find on patient data that physicians/coders at different hospitals sometimes diagnose/record different conditions in the electronic medical record. However, the majority of patient data is similarly diagnosed and recorded across hospitals (motivating the common practice of borrowing risk predictors from other hospitals). This creates a bias between the proxy and true predictive tasks that can be captured by the few features corresponding only to the subset of diagnoses where differences arise. 

Importantly, in both examples, the proxy and gold estimators themselves are \textit{not} sparse. Thus, we cannot exploit this structure by directly applying high-dimensional techniques to proxy or gold data separately. Rather, we must efficiently combine proxy and gold data, while exploiting the sparse structure of the bias \textit{between} the two predictive tasks.
Our lower bounds show that popular heuristics (model averaging and weighted loss functions) fail to leverage sparse structure even when it is present, and can still only improve predictive accuracy by at most a constant factor.

We propose a new two-step joint estimator that successfully leverages sparse structure in the bias term to achieve a much stronger improvement in predictive accuracy. In particular, our proposed estimator can achieve the same accuracy with \textit{exponentially} less gold data (in the number of features $d$). Intuitively, instead of using the limited gold data directly for estimating the predictive model, our estimator uses gold data to efficiently de-bias the proxy estimator. In fact, when gold data is very limited, the availability of proxy data is critical to extracting value from the gold data. Our proof relies on a new LASSO tail inequality for approximately sparse vectors, which may be of independent interest. It is worth noting that our estimator does not simultaneously require both proxy and gold data at training time; this is an important feature in settings such as healthcare, where data from different sources cannot be combined due to regulatory constraints. We demonstrate the effectiveness of our estimator on both Expedia hotel recommendation (Example \ref{ex:expedia}) and diabetes risk prediction (Example \ref{ex:diabetes}). In both cases, we achieve significantly better predictive accuracy, as well as managerial insights into the nature of the bias in the proxy data.

\subsection{Pervasiveness of Proxies} \label{ssec: more-proxies}

Proxies are especially pervasive in healthcare, where patient covariates and response variables must be derived from electronic medical records (EMRs), which are inevitably biased by the data collection process. One common issue is \textit{censoring}: we only observe a diagnosis in the EMR \textit{if} the patient visits the healthcare provider. Thus, the recorded diagnosis code (often used as the response variable) is in fact a proxy for the patient's true outcome (which may or may not have been recorded). \cite{mullainathan2017does} and \cite{obermeyer2017lost} demonstrate that this proxy can result in misleading predictive models, arising from systematic biases in the types of patients who frequently visit the healthcare provider. One could collect more reliable (true) outcome data by surveying patients, but this is costly and only scales to a small cohort of patients.
Another form of censoring is \textit{omitted variable bias}: important factors (e.g., physician counseling or a patient's proactiveness towards their own health) are not explicitly recorded in the medical record.  \cite{bastani2017} show that omitted variable bias arising from unrecorded physician interventions can lead to misleading predictive models trained on EMR data. Again, more reliable (gold) data can be collected by hand-labeling patient observations based on physician or nurse notes in the medical chart, but as before, this is costly and unscalable.
Recently, researchers have drawn attention to \textit{human bias}: patient data is collected and recorded by hospital staff (e.g., physicians, medical coders), who may themselves be biased \citep{ahsen2018algorithmic}. This is exemplified in our case study (Section \ref{ssec:diabetes}), where we find that medical coders record the obesity diagnosis code in the EMR at very different rates even when patient BMIs are similar.
Finally, the specific outcomes of interest may be \textit{too rare} or have high variance. For example, in healthcare pay-for-performance contracts, Medicare uses 30-day hospital readmissions rates as proxies for hospital quality of care, which may be better captured by rarer outcomes such as never events or 30-day patient mortality rates \citep{CMS, axon2011hospital, milstein2009ending}.

Proxies are also pervasive in marketing and revenue management. Online platforms allow us to observe fine-grained customer behaviors, including page views, clicks, cart-adds, and eventually purchases. While purchases may be the final outcome of interest, these intermediate (and more abundant) observations serve as valuable proxies. For example, \cite{farias2017learning} use a variety of customer actions as proxies for predicting a customer's affinity for a song in a music streaming service. This is also evidenced in our case study (Section \ref{ssec:expedia}), where customer clicks can signal the likelihood of customer hotel purchases. With modern technology, companies can also observe customers' offline behavior, including store visits (using mobile WiFi signal tracking, e.g., see \citealp{zhang2018value} for Alibaba case study) and real-time product browsing (using store security cameras, e.g., see \citealp{brynjolfsson2013competing} for American Apparel case study). Thus, different channels of customer behavior can inform predictive analytics. For example, \cite{dzyabura2018accounting} use online customer behaviors as proxies for predicting offline customer preferences. Finally, new product introduction can benefit from proxies. For example, \cite{baardman2017leveraging} use demand for related products as proxies for predicting demand for a new product.

\subsection{Other Related Work}

Our problem can be viewed as an instance of multitask learning, or more specifically, \textit{transfer learning}.
Multitask learning combines data from multiple related predictive tasks to train similar predictive models for each task. It does this by using a \textit{shared representation} across tasks \citep{caruana1997multitask}. Such representations typically include variable selection (i.e., enforce the same feature support for all tasks in linear or logistic regression, \citealp{jalali2010dirty, meier2008group}), kernel choice (i.e., use the same kernel for all tasks in kernel regression, \citealp{caruana1997multitask}), or intermediate neural net representations (i.e., use the same weights for intermediate layers for all tasks in deep learning, \citealp{collobert2008unified}). Transfer learning specifically focuses on learning a single new task by transferring knowledge from a related task that has already been learned (see \citealp{pan2010survey} for a survey). We share a similar goal: since we have many proxy samples, we can easily learn a high-performing predictive model for the proxy task, but we wish to transfer this knowledge to the (related) gold task for which we have very limited labeled data. However, our proxy and gold predictive models already have a shared representation in the variable selection sense; in particular, we use the same features (all of which are typically relevant) for both prediction tasks. 

We note that the tasks considered in the multitask and transfer learning literature are typically far more disparate than the class of proxy problems we have identified in this paper thus far. For instance, \cite{caruana1997multitask} gives the example of simultaneously training neural network outputs to recognize different object properties (outlines, shapes, textures, reflections, shadows, text, orientation, etc.). \cite{bayati2018statistical} simultaneously train logistic regressions predicting disparate diseases (heart failure, diabetes, dementia, cancer, pulmonary disorder, etc.). While these tasks are indeed related, they are not close substitutes for each other. In contrast, the proxy predictive task \textit{is} a close substitute for the true predictive task, to the point that practitioners may even ignore gold data and train their models purely on proxy data. In this class of problems, we can impose significantly more structure beyond merely a shared representation.

Our key insight is that the bias between the proxy and gold predictive tasks can be modeled as a sparse function. We argue that there is often some (a priori unknown) underlying mechanism that systematically affects a subset of the features, creating a bias between the true and proxy predictive tasks. When this is the case, we can successfully estimate the bias using high-dimensional techniques that exploit sparsity.

Bayesian approaches have been proposed for similar problems. For instance, \cite{dzyabura2018accounting} use a Bayesian prior relating customers' online preferences (proxies) and offline purchase behavior (true outcome of interest). \cite{raina2006constructing} propose a method for constructing priors in such settings using semidefinite programming on data from related tasks. These approaches do not come with theoretical convergence guarantees. A frequentist interpretation of their approach is akin to ridge regression, which is one of our baselines; we prove that ridge regression cannot take advantage of sparse structure when present, and thus, cannot significantly improve improve estimation error over the naive proxy or gold estimators. Relatedly, \cite{farias2017learning} link multiple low-rank collaborative filtering problems by imposing structure across their latent feature representations; however, the primary focus in their work is on low-rank matrix completion settings without features, whereas our focus is on classical regression problems.

We use techniques from the high-dimensional statistics literature to prove convergence properties about our two-step estimator. The second step of our estimator uses a LASSO regression \citep{donoho, tibshirani}, which helps us recover the bias term using far fewer samples than traditional statistical models by exploiting sparsity \citep{candes, bickel, negahban}. A key challenge in our proof is that the vector we wish to recover in the second stage is not perfectly sparse; rather, it is the sum of a sparse vector and residual noise from the first stage of our estimator. Existing work has studied convergence of LASSO for approximately sparse vectors \citep{buhlmann, belloni2012sparse}, while making little to no assumptions on the nature of the approximation; we extend this theory to prove a tighter but structure-dependent tail inequality that is appropriate for our setting. As a result, we show that the error of our joint estimator cleanly decomposes into a term that is proportional to the variance of our proxy estimator (which is small in practice), and a term that recovers the classical error rate of the LASSO estimator. Thus, when we have many proxy observations, we require exponentially fewer gold observations to achieve a nontrivial estimation error than would be required if we did not have any proxy data. Our two-stage estimator is related in spirit to other high-dimensional two-stage estimators \citep[e.g.,][]{belloni, belloni2012sparse}. While these papers focus on treatment effect estimation after variable selection on features or instrumental variables, our work focuses on transfer learning from a proxy predictive task to a new predictive task with limited labeled data.

\subsection{Contributions}
We highlight our main contributions below:
\begin{enumerate}
\item \textit{Problem Formulation:} We formulate the proxy problem as two classical regression tasks; the proxy task has abundant data, while the actual (gold) task of interest has limited data. Motivated by real datasets, we model the bias between the two tasks as a sparse function of the features.
\item \textit{Theory:} We propose a new two-step estimator that efficiently combines proxy and gold data to exploit sparsity in the bias term. Our estimator provably achieves the same accuracy as popular heuristics (e.g., model averaging or weighted loss functions) with exponentially less gold data (in the number of features $d$). Our proof relies on a new tail inequality on the convergence of LASSO for approximately sparse vectors, which may be of independent interest.
\item \textit{Case Studies:} We demonstrate the effectiveness of our approach on e-commerce and healthcare datasets. In both cases, we achieve significantly better predictive accuracy as well as managerial insights into the nature of the bias in the proxy data.
\end{enumerate}

\section{Problem Formulation} \label{sec:formulation}

\textit{Preliminaries:} For any integer $n$, let $[n]$ denote the set $\{1,...,n\}$. Consider an observation with feature vector $\vx \in \R^d$. As discussed earlier, the gold and predictive tasks are different. Let the gold and proxy responses be given by the following linear data-generating processes respectively (we will discuss nonlinear parametric models in Section \ref{ssec:nonlinear}):
\begin{align*}
\yg &= \vx^\top \betag + \epg \,, \\
\yp &= \vx^\top \betap + \epp \,,
\end{align*}
where $\betag, \betap \in \R^d$ are unknown regression parameters, and the noise $\epg, \epp$ are each vectors of independent subgaussian variables with parameters $\sigmag$ and $\sigmap$ respectively (see Definition \ref{def:subgaussian} below). We do not impose that $\epg$ is independent of $\epp$; for example, $(\epg^{(i)}, \epp^{(i)})$ can be arbitrarily pairwise correlated\footnote{In Example \ref{ex:expedia}, a customer's click (proxy) and purchase (gold) responses may have correlated customer-specific noise.} for any $i$. 

\begin{definition}\label{def:subgaussian}
A random variable $z \in \R$ is $\sigma$-subgaussian if $\E[e^{tz}] \leq e^{\sigma^2 t^2/2} $ for every $t \in \R$.
\end{definition}
This definition implies $\E[z] = 0$ and $\text{Var}[z] \leq \sigma^2$. Many classical distributions are subgaussian; typical examples include any bounded, centered distribution, or the normal distribution. Note that the errors need not be identically distributed. 

Our goal is to estimate $\betag$ accurately in order to make good decisions for new observations with respect to their true predicted outcomes. In a typical regression problem, the gold data would suffice. However, we often have very limited gold data, leading to high-variance erroneous estimates. This can be either because $\ngold$ is small (Example \ref{ex:diabetes}) or $\sigmag$ is large (Example \ref{ex:expedia}). Thus, we can benefit by utilizing information from proxy data, even if this data is biased.

Decision-makers employ proxy data because the proxy predictive task is closely related to the true predictive task. In other words, $\betag \approx  \betap$. To model the relationship between the true and proxy predictive tasks, we write
\[ \betag = \betap + \delta^* \,,\]
where $\delta^*$ captures the proxy estimator's bias.

Motivated by our earlier discussion, we posit that the bias is \textit{sparse}. In particular, let $\| \delta^* \|_0 = s$, which implies that the bias of the proxy estimator only depends on $s$ out of the $d$ covariates. This constraint is always satisfied when $s=d$, but we will prove that our estimator of $\betag$ has much stronger performance guarantees when $s \ll d$.

\textit{Data:} We are given two (possibly overlapping) cohorts. We have $\ngold$ observations in our gold dataset: let $\Xg \in \R^{\ngold \times d}$ be the gold design matrix (whose rows are observations from the gold cohort), and $\Yg \in \R^{\ngold}$ be the corresponding vector of responses. Analogously, we have $\nproxy$ observations in our proxy dataset: let $\Xp \in \R^{\nproxy \times d}$ be the proxy design matrix (whose rows are observations from the proxy cohort), and $\Yp \in \R^{\nproxy}$ be the corresponding vector of responses. Typically $\ngold \ll \nproxy$ or $\sigmag \gg \sigmap$, necessitating the use of proxy data. Without loss of generality, we impose that both design matrices have been standardized, i.e., 
\[ \bn{\Xg^{(r)}}_2^2 = \ngold \text{~~~ and ~~~} \bn{\Xp^{(r)}}_2^2 = \nproxy \,,\]
for each column $r \in [d]$. It is standard good practice to normalize features in this way when using regularized regression, so that the regression parameters are appropriately scaled in the regularization term \citep[see, e.g.,][]{friedman2001}.
We further define the $d\times d$ gold and proxy \textit{sample covariance matrices}
\begin{align*}
\Sigmag = \frac{1}{\ngold}\Xg^\top \Xg  \text{~~~ and ~~~} 
\Sigmap = \frac{1}{\nproxy} \Xp^\top \Xp \,.
\end{align*}
Our standardization of the design matrices implies that 
$\text{diag}\bp{\Sigmag} = \text{diag}\bp{\Sigmap} = \mathbf{1}_{d\times 1}$.

\textit{Evaluation:} We define the parameter estimation error of a given estimator $\hbeta$ relative to the true parameter $\betag$ as
\[ R(\hbeta, \betag) = \sup_{\mathcal{S}} ~\E\bb{\bn{\hbeta - \betag}_1} \,,\]
where $\mathcal{S} = \bc{\Xg, \Xp, \betag, \delta^*}$ is the set of feasible problem parameters\footnote{Note that $\betap$ is implicitly defined in $\mathcal{S}$ as $\betag - \delta^*$.} (i.e., satisfying the assumptions given in the problem formulation and Section \ref{ssec:ass}), and the expectation is taken with respect to the noise terms $\epg$ and $\epp$. Note that a bound on $R$ implies a bound on the expected out-of-sample prediction error for any new bounded observation $x \in \R^d$, i.e., by H\"older's inequality,
\[ \E\bb{\left|x^\top\hbeta - x^\top\betag\right|} ~\leq~ \E\bb{\bn{\hbeta - \betag}_1} \cdot \bn{x}_\infty ~\leq~ R(\hbeta, \betag) \cdot \bn{x}_\infty \,.\]

\subsection{Assumptions} \label{ssec:ass}

\begin{assumption}[Bounded] \label{ass:bounded}
There exists some $b \in \R$ such that $\bn{\betag}_1 \leq b$.
\end{assumption}

Our first assumption states that our regression parameters are bounded by some constant\footnote{Note that this does not imply that $\betag$ is sparse, e.g., $u = \frac{1}{d}\mathbf{1} \in \R^d$ satisfies $\bn{u}_1 = 1$ but $\bn{u}_0 = d$.}. This is a standard assumption in the statistical literature.

\begin{assumption}[Positive-Definite] \label{ass:proxy-posdef} The proxy sample covariance matrix $\Sigmap$ is positive-definite. In other words, the minimum eigenvalue of $\Sigmap$ is $\psi > 0$.
\end{assumption}

Our second assumption is also standard, and ensures that $\betap$ is identifiable from the proxy data $(\Xp, \Yp)$. This is a mild assumption since $\nproxy$ is large. In contrast, we allow that $\betag$ may not be identifiable from the gold data $(\Xg, \Yg)$, since $\ngold$ is small and the resulting sample covariance matrix $\Sigmag$ may not be positive-definite.

The last assumption on the \textit{compatibility condition} arises from the theory of high-dimensional statistics \citep{candes,bickel,buhlmann}. We will require a few definitions before stating the assumption.

An \textit{index set} is a set $S \subset[d]$. For any vector $u \in\R^d$, let $u_{S}\in\R^{d}$ be the vector obtained by setting the elements of $u$ that are not in $S$ to zero. Then, the $i^{th}$ element of $u_{S}$ is $u_{S}^{(i)} = u^{(i)} \cdot \mathbbm{1}[i \in S]$. Furthermore, let $S^c$ denote the complement of $S$. Then, $S \cup S^c = [d]$ and $S \cap S^c = \varnothing$.

The \textit{support} for any vector $u \in \R^d$, denoted $supp(u) \subset [d]$, is the set of indices corresponding to nonzero entries of $u$. Thus, $supp(u)$ is the smallest set that satisfies $u_{supp(u)} = u$.

We now define the compatibility condition:
\begin{definition}[Compatibility Condition] \label{def:cc} The compatibility condition is met for the index set $S \subseteq [d]$ and the matrix $\Sigma \in \R^{d \times d}$ if there exists $\phi > 0$ such that, for all $u \in \R^d$ satisfying $\| u_{S^c} \|_1 \leq 3 \|u_S\|_1$, it holds that
\[ \|u_{S}\|_1^2 \leq \frac{|S|}{\phi^2}\bp{u^T \Sigma u} \,. \]
\end{definition}

\begin{assumption}[Compatibility Condition] \label{ass:cc} The compatibility condition (Definition \ref{def:cc}) is met for the index set $S = supp(\delta^*)$ and gold sample covariance matrix $\Sigmag$ with constant $\phi > 0$.
\end{assumption}

Our third assumption is critical to ensure that the bias term $\delta^*$ is identifiable, even if $\ngold < d$. This assumption (or the related restricted eigenvalue condition) is standard in the literature to ensure the convergence of high-dimensional estimators such as the Dantzig selector or LASSO \citep{candes, bickel, buhlmann}. 

It is worth noting that Assumption \ref{ass:cc} is always satisfied if $\Sigmag$ is positive-definite. In particular, letting $\zeta > 0$ be the minimum eigenvalue of $\Sigmag$, it can be easily verified that the compatibility condition holds with constant $\phi_0 = \sqrt{\zeta}$ for \textit{any} index set. Thus, the compatibility condition is strictly weaker than the requirement that $\Sigmag$ be positive-definite. For example, the compatibility condition allows for collinearity in features that are outside the index set $S$, which can occur often in high-dimensional settings when $|S| = s \ll d$ \citep{buhlmann}. Thus, even when $\betag$ is not identifiable, we may be able to identify the bias $\delta^*$ by exploiting sparsity.

\section{Baseline Estimators} \label{sec:baselines}

We begin by describing four commonly used baseline estimators. These include naive estimators trained only on gold or proxy data, as well as two popular heuristics (model averaging and weighted loss functions). We prove corresponding lower bounds on their parameter estimation error $R(\cdot, \betag)$ with respect to the true parameter $\betag$.

\subsection{OLS/Ridge Estimator on Gold Data}

One common approach is to ignore proxy data and simply use the gold data (the most appropriate data) to construct the best possible predictor. Since we have a linear model, the ordinary least squares (OLS) estimator is the most obvious choice: it is the minimum variance unbiased estimator.

However, it is well known that introducing bias can be beneficial in data-poor environments. In other words, since we have very few gold samples ($\ngold$ is small), we may wish to consider the regularized ridge estimator \citep{friedman2001}:
\[ \hbetag^{ridge}(\lambda) ~=~ \arg\min_{\beta} \bc{ \frac{1}{\ngold} \bn{ \Yg - \Xg \beta }_2^2 + \lambda \bn{\beta}_2^2} \,,\]
where we introduce a regularization parameter $\lambda \geq 0$. Note that when the regularization parameter $\lambda = 0$, we recover the classical OLS estimator, i.e., $\hbetag^{ridge}(0) = \hbetag^{OLS}$.

\begin{theorem}[Gold Estimator] \label{thm:gold-ridge}
The parameter estimation error of the OLS estimator on gold data $\bp{\Xg, \Yg}$ is bounded below as follows:
\begin{align*}
R\bp{  \hbetag^{OLS}, \betag } ~&\geq~ d \sqrt{\frac{2\sigmag^2}{\pi \ngold}} \\
&=~ \cO\bp{\frac{d\sigmag}{\sqrt{\ngold}}}\,.
\end{align*} 
The parameter estimation error of the ridge estimator on gold data $\bp{\Xg, \Yg}$ for any choice of the regularization parameter $\lambda \geq 0$ is bounded below as follows:
\begin{align*}
\min_{\lambda \geq 0}~R\bp{ \hbetag^{ridge}(\lambda), \betag } ~&\geq~ \frac{ d\sigmag /\sqrt{2\pi}}{b\sqrt{\ngold} + d\sigmag \sqrt{2/\pi}} \\
&=~ \cO\bp{\frac{ d\sigmag}{\sqrt{\ngold} + d\sigmag}} \,.
\end{align*}
\end{theorem}
The proof is given in Appendix \ref{app:base-gold}. Note that this result uses the optimal value of the regularization parameter $\lambda$ to compute the lower bound on the parameter estimation error of the ridge estimator. In practice, the error will be larger since $\lambda$ would be estimated through cross-validation.

Theorem \ref{thm:gold-ridge} shows that when the number of gold samples is moderate (i.e., $\ngold \gg d\sigmag$), the ridge estimator recovers the OLS estimator's lower bound on the parameter estimation error $\mathcal{O}\bp{d \sigmag / \ngold}$. However, when the number of gold samples is very small (i.e., $\ngold \lesssim d\sigmag$), the ridge estimator achieves a constant lower bound on the parameter estimation error $\mathcal{O}(1)$. This is because the ridge estimator will predict $\hbetag^{ridge}(\lambda) = 0$ for very small values of $\ngold$, and since we have assumed that $\bn{\betag}_1 \leq b = \mathcal{O}(1)$, our parameter estimation error remains bounded.

\subsection{OLS Estimator of Proxy Data}

Another common approach is to ignore the gold data and simple use the proxy data to construct the best possible predictor. Since we have a linear model, the OLS estimator is the most obvious choice; note that we do not need regularization since we have many proxy samples ($\nproxy$ is large). Thus, we consider:
\[ \hbetap ~=~ \arg\min_{\beta} \bc{ \frac{1}{\nproxy} \| \yp - \Xp \beta \|_2^2} \,. \]

\begin{theorem}[Proxy Estimator] \label{thm:proxy-ols}
The parameter estimation error of the OLS estimator on proxy data $\bp{\Xp, \Yp}$ is bounded below as follows:
\begin{align*}
R\bp{  \hbetap, \betag } ~&\geq~ \max \bc{ \frac{1}{2} \bn{\delta^*}_1, d\sqrt{\frac{\sigmap^2}{2\pi \nproxy}}} \\
&=~ \cO\bp{\bn{\delta^*}_1 + \frac{d\sigmap}{\sqrt{\nproxy}}}\,.
\end{align*} 
\end{theorem}
The proof is given in Appendix \ref{app:base-proxy}. Since $\nproxy$ is large, the second term in the parameter estimation error $d\sigmap/\sqrt{\nproxy}$ is small. Thus, the parameter estimation error of the proxy estimator is dominated by the bias term $\bn{\delta^*}_1$. When the proxy is  ``good" or reasonably representative of the gold data, $\bn{\delta^*}_1$ is small. In these cases, the proxy estimator is more accurate than the gold estimator, explaining the widespread use of the proxy estimator in practice even when (limited) gold data is available.

\subsection{Model Averaging Estimator} \label{ssec:avg}

One heuristic that is sometimes employed is to simply average the gold and proxy estimators:
\[ \hbeta_{avg}(\lambda) ~=~ (1-\lambda)\cdot \hbetag^{OLS} + \lambda \cdot \hbetap \,, \]
for some averaging parameter $\lambda \in [0,1]$. Note that $\lambda=0$ recovers $\hbetag^{OLS}$ (the OLS estimator on gold data) and $\lambda=1$ recovers $\hbetap$ (the OLS estimator on proxy data).

\begin{theorem}[Averaging Estimator] \label{thm:avg}
The parameter estimation error of the averaging estimator on both gold and proxy data $\bp{\Xg, \Yg, \Xp, \Yp}$ is bounded below as follows:
\begin{align*}
\min_{\lambda \in [0,1]} ~R\bp{\hbeta_{avg}(\lambda), \betag } ~&\geq~ \min\bc{ \frac{d \sigmag}{3\sqrt{2\pi \ngold}}, ~\frac{1}{6} \bn{\delta^*}_1 + \frac{d \sigmap}{3\sqrt{ 2\pi \nproxy}}  } \\
&=~ \cO\bp{\min\bc{ \frac{d \sigmag}{\sqrt{\ngold}},~\bn{\delta^*}_1 + \frac{d \sigmap}{\sqrt{ \nproxy}}  }} \,.
\end{align*}
\end{theorem}
The proof is given in Appendix \ref{app:base-avg}. Note that this result uses the optimal value of the averaging parameter $\lambda$ to compute the lower bound on the parameter estimation error of the averaging estimator. In practice, the error will be larger since $\lambda$ would be estimated through cross-validation.

Theorem \ref{thm:avg} shows that the averaging estimator does not achieve more than a constant factor improvement over the best of the gold and proxy OLS estimators. In particular, the lower bound in Theorem \ref{thm:avg} is exactly the minimum of the lower bounds of the gold OLS estimator (given in Theorem \ref{thm:gold-ridge}) and the proxy OLS estimator (given in Theorem \ref{thm:proxy-ols}) up to constant factors. Since the averaging estimator spans both the proxy and the gold estimators (depending on the choice of $\lambda$), it is to be expected that the best possible averaging estimator does at least as well as either of these two estimators; surprisingly, it does no better.

\subsection{Weighted Loss Estimator} A more sophisticated heuristic used in practice is to perform a weighted regression that combines both datasets but assigns a higher weight to true outcomes. Consider:
\[ \hbeta_{weight}(\lambda) ~=~ \arg\min_{\beta} \bc{ \frac{1}{\lambda \ngold + \nproxy} \cdot \bp{ \lambda \| \Yg - \Xg \beta \|_2^2 + \| \Yp - \Xp \beta \|_2^2 }} \,, \]
for some weight $\lambda \in [0, \infty)$. Note that $\lambda =\infty$ recovers $\hbetag^{OLS}$ (the OLS estimator on gold data) and $\lambda = 0$ recovers $\hbetap$ (the OLS estimator on proxy data).

\begin{theorem}[Weighted Loss Estimator] \label{thm:weighted}
The parameter estimation error of the weighted estimator on both gold and proxy data $\bp{\Xg, \Yg, \Xp, \Yp}$ is bounded below as follows:
\begin{align*}
\min_{\lambda\geq 0}~R \bp{ \hbeta_{weight}(\lambda), \betag } ~&\geq~ \min\bc{ \frac{d \sigmag}{3\sqrt{2\pi \ngold}}, ~\frac{1}{6} \bn{\delta^*}_1 + \frac{d \sigmap}{3\sqrt{ 2\pi \nproxy}}  } \\
&=~ \cO\bp{\min\bc{ \frac{d \sigmag}{\sqrt{\ngold}},~\bn{\delta^*}_1 + \frac{d \sigmap}{\sqrt{ \nproxy}}  }} \,.
\end{align*}
\end{theorem}
The proof is given in Appendix \ref{app:base-wt}. Note that this result uses the optimal value of the weighting parameter $\lambda$ to compute the lower bound on the parameter estimation error of the weighted loss estimator. In practice, the error will be larger since $\lambda$ would be estimated through cross-validation.

Theorem \ref{thm:weighted} shows that the more sophisticated weighted loss estimator achieves exactly the same lower bound as the averaging estimator (Theorem \ref{thm:avg}). Thus, the weighted loss estimator also does not achieve more than a constant factor improvement over the best of the gold and proxy estimators. Since the weighted estimator spans both the proxy and the gold estimators (depending on the choice of $\lambda$), it is to be expected that the best possible weighted estimator does at least as well as either of these two estimators; again, surprisingly, it does no better.

As discussed earlier, prediction error is composed of bias and variance. Training our estimator on the true outcomes alone yields an unbiased but high-variance estimator. On the other hand, training our estimator on the proxy outcomes alone yields a biased but low-variance estimator. Averaging the estimators or using a weighted loss function can interpolate the bias-variance tradeoff between these two extremes, but provides at most a constant improvement in prediction error.

\section{Joint Estimator}

We now define our proposed joint estimator, and prove that it can leverage sparsity to achieve much better theoretical guarantees than common approaches used in practice.

\subsection{Definition} \label{ssec:joint-def}

We propose the following two-step joint estimator $\hbeta_{joint}(\lambda)$:
\begin{align}
\textbf{Step 1: ~~~}& \hbeta_{proxy} = \arg\min_{\beta}~ \bc{ \frac{1}{\nproxy} \bn{\Yp - \Xp \beta }^2_2 } \nonumber \\
\textbf{Step 2: ~~~}& \hbeta_{joint}(\lambda) = \arg\min_{\beta}~ \bc{ \frac{1}{\ngold} \bn{\Yg - \Xg \beta }^2_2 + \lambda\bn{\beta - \hbeta_{proxy} }_1 } \,. \label{eq:obj}
\end{align}
Both estimation steps are convex in $\beta$. Thus, there are no local minima, and we can find the global minimum through standard techniques such as stochastic gradient descent. Note that the first step only requires proxy data, while the second step only requires gold data; thus, we do not need both gold and proxy data to be simultaneously available during training. This is useful when data from multiple sources cannot be easily combined, but summary information like $\hbetap$ can be shared.

When the regularization parameter $\lambda$ is small, we recover the gold OLS estimator; when $\lambda$ is large, we recover the proxy OLS estimator. Thus, similar to model averaging and weighted loss functions, the joint estimator spans both the proxy and the gold estimators (depending on the choice of $\lambda$). However, we show that the joint estimator can successfully interpolate the bias-variance tradeoff between these extremes to produce up to an exponential reduction in estimation error.

Intuitively, we seek to do better by leveraging our insight that the bias term $\delta^*$ is well-modeled by a sparse function of the covariates. Thus, in principle, we can efficiently recover $\delta^*$ using an $\ell_1$ penalty. A simple variable transformation of the second-stage objective (\ref{eq:obj}) gives us
\begin{equation}
\hdelta(\lambda) = \arg\min_\delta \bc{ \frac{1}{\ngold} \bn{ \Yg - \Xg ( \delta + \hbeta_{proxy}) }_2^2 + \lambda \| \delta \|_1} \,,  \label{eq:obj2}
\end{equation}
where we have taken $\delta = \beta - \hbetap$. Our estimator is then simply $\hbeta_{joint}(\lambda) = \hdelta(\lambda) + \hbetap$, where $\hbetap$ is estimated in the first stage. In other words, (\ref{eq:obj2}) uses the LASSO estimator on gold data to recover the bias term with respect to the proxy estimator $\hbetap$. We use the $\ell_1$ penalty, which is known to be effective at recovering sparse vectors \citep{candes}. 

This logic immediately indicates a problem, because the parameter we wish to converge to in (\ref{eq:obj2}) is not actually the sparse vector $\delta^*$, but a combination of $\delta^*$ and residual noise from the first stage. 
We formalize this by defining some additional notation:
\begin{align}
\nu &~=~  \hbetap - \betap  \,, \\
\tdelta &~=~ \beta_{gold}^* -\hbeta_{proxy} ~=~ \delta^* - \nu \,.
\end{align}
Here, $\nu$ is the residual noise in estimating the proxy estimator $\hbetap$ from the first stage. As a consequence of this noise, in order to recover the true gold parameter $\betag = \tdelta + \hbetap$, we wish to recover $\tdelta$ (rather than $\delta^*$) from (\ref{eq:obj2}). Specifically, note that the minimizer of the first term in (\ref{eq:obj2}) is $\tdelta$ and not $\delta^*$. However, $\tdelta$ is clearly not sparse, since $\nu$ is not sparse (e.g., if the noise $\epp$ is a gaussian random variable, then $\nu$ is also gaussian). Thus, we may be concerned that the LASSO penalty in (\ref{eq:obj2}) may not be able to recover $\tdelta$ at the exponentially improved rate promised for sparse vectors \citep{candes,bickel,buhlmann}.

On the other hand, since we have many proxy outcomes ($\nproxy$ is large), our proxy estimation error $\bn{\nu}_1$ is small. In other words, $\tdelta$ is \textit{approximately} sparse. We will prove that this is sufficient for us to recover $\tdelta$ (and therefore $\beta_{gold}^*$) at an exponentially improved rate.

\subsection{Main Result}

We now state a tail inequality that upper bounds the parameter estimation error of the two-step joint estimator with high probability.
\begin{theorem}[Joint Estimator] \label{thm:main} The joint estimator satisfies the following tail inequality for any chosen value of the regularization parameter $\lambda > 0$:
\[ \Pr\bb{\bn{ \hbeta_{joint}(\lambda) - \betag  }_1 ~\geq~ 5\lambda \bp{\frac{1}{4\psi^2} + \frac{1}{\psi} + \frac{s}{ 2\phi^2} }} ~\leq~ 2d  \exp\bp{-\frac{\lambda^2 \ngold}{200 \sigmag^2}} + 2d\exp\bp{-\frac{\lambda^2 \nproxy}{2 d^2 \sigmap^2}}  \,.\]
\end{theorem}
The proof is given in Section \ref{ssec:proof} with supporting lemmas in Appendix \ref{app:joint-linear}. The regularization parameter trades off the bound on the parameter estimation error $\bn{ \hbeta_{joint}(\lambda) - \betag  }_1$ with the probability that the bound holds. If $\lambda$ is too small, the guarantee in Theorem \ref{thm:main} becomes trivial, since the probability of deviation is upper bounded by $1$. Thus, $\lambda$ must be chosen appropriately to achieve a reasonable bound on the parameter estimation error with relatively high probability. 

In a typical LASSO problem, an optimal choice of the regularization parameter is $\lambda = \tO\bp{\sigmag/\sqrt{\ngold}}$. However, in Theorem \ref{thm:main}, convergence depends on both gold \textit{and} proxy data. In Corollary \ref{cor:joint}, we will show that in this setting, we will need to choose 
\[ \lambda = \tO\bp{\frac{\sigmag}{\sqrt{\ngold}} + \frac{d\sigmap}{\sqrt{\nproxy}}} \,. \]
In the next subsection, we will compute the resulting estimation error of the joint estimator.

\begin{remark}
We can extend our approach to prove bounds on the $\ell_2$ estimation error $\bn{\hbeta_{joint} - \betag}_2$ by replacing our compatibility condition (Assumption \ref{ass:cc}) with a restricted eigenvalue condition \citep{candes, bickel}. This extension is provided in Appendix \ref{app:l2}, and consistent with the literature, yields bounds that scale as $\sqrt{s}$ rather than $s$.
\end{remark}

\subsection{Comparison with Baselines}

We now derive an upper bound on the expected parameter estimation error of the joint estimator, in order to compare its performance against the baseline estimators described in Section \ref{sec:baselines}.

From Theorem \ref{thm:main}, we know that our estimation error $\bn{ \hbeta_{joint} - \betag }_1$ is small with high probability. However, to derive an upper bound on $R(\cdot)$, we also need to characterize its worst-case magnitude. In order to ensure that our estimator $\hbeta_{joint}$ never becomes unbounded, we consider the \textit{truncated} joint estimator $\hbeta_{joint}^{tr}$. In particular, 
\[ \hbeta_{joint}^{tr}=
\begin{cases}
    \hbeta_{joint} & \text{if $\bn{\hbeta_{joint} }_1 \leq 2b$} \,, \\
    0 & \text{otherwise}\,.
  \end{cases}
\]
Recall that $b$ is any upper bound on $\bn{\betag}_1$ (Assumption \ref{ass:bounded}), and can simply be considered a large constant.
The following corollary uses the tail inequality in Theorem \ref{thm:main} to obtain an upper bound on the expected parameter estimation error of the truncated joint estimator.
\begin{corollary}[Joint Estimator] \label{cor:joint} The parameter estimation error of the truncated joint estimator on both gold and proxy data $\bp{\Xg, \Yg, \Xp, \Yp}$ is bounded above as follows:
\begin{align*}
R \bp{ \hbeta_{joint}^{tr}(\lambda), \betag } ~&\leq~ 5\lambda \bp{\frac{1}{4\psi^2} + \frac{1}{\psi} + \frac{s}{ 2\phi^2} } + 6bd \bp{\exp\bp{-\frac{\lambda^2 \ngold}{200 \sigmag^2}} + \exp\bp{-\frac{\lambda^2 \nproxy}{2 d^2 \sigmap^2}}}\,.
\end{align*}
Let $C> 0$ be any tuning constant. Taking the regularization parameter to be
\begin{align*}
\blambda ~&=~ C\max\bc{\sqrt{\frac{200 \sigmag^2 \log\bp{6bd\ngold}}{\ngold}},~ \sqrt{\frac{2d^2 \sigmap^2 \log\bp{6bd\nproxy}}{\nproxy}}} =~ \tO\bp{\frac{\sigmag }{\sqrt{\ngold}} + \frac{d \sigmap }{ \sqrt{\nproxy}}} \,,
\end{align*}
yields a parameter estimation error of order
\begin{align*}
R \bp{ \hbeta_{joint}^{tr}(\blambda), \betag } ~&=~ \cO\bp{\max\bc{ \frac{s \sigmag }{\sqrt{\ngold}} \log\bp{d\ngold} ,~ \frac{sd \sigmap }{\sqrt{\nproxy}}\log\bp{d\nproxy}}} \,.
\end{align*}
\end{corollary}
The proof is given in Appendix \ref{app:corollary}. The parameter estimation error $R \bp{ \hbeta_{joint}^{tr}(\blambda), \betag }$ cleanly decomposes into two terms: (i) the first term is the classical error rate of the LASSO estimator \textit{if} $\betag$ (rather than $\delta^*$) was sparse, and (ii) the second term is proportional to the error of the proxy estimator \textit{if} there were no bias (i.e., $\delta^* = 0$).

\begin{table}[h]
\centering
\begin{tabular}{@{}lcc@{}}
\toprule
\textbf{Estimator} & \begin{tabular}[c]{@{}c@{}}\textbf{Parameter Estimation Error}\\ \textit{(up to constants)}\end{tabular} & \textbf{Bound Type} \\ \midrule \addlinespace[0.5em]
Gold OLS & $\dfrac{d\sigmag}{\sqrt{\ngold}}$ & Lower \\
\addlinespace[0.5em]
Gold Ridge & $\dfrac{d\sigmag}{\sqrt{\ngold} + d\sigmag}$ & Lower \\
\addlinespace[0.5em]
Proxy OLS & $\bn{\delta^*}_1 + \dfrac{d\sigmap}{\sqrt{\nproxy}}$ & Lower \\
\addlinespace[0.5em]
Averaging & $\min\bc{\dfrac{d\sigmag}{\sqrt{\ngold}},~ \bn{\delta^*}_1 + \dfrac{d\sigmap}{\sqrt{\nproxy}}}$ & Lower \\
\addlinespace[0.5em]
Weighted & $\min\bc{\dfrac{d\sigmag}{\sqrt{\ngold}},~ \bn{\delta^*}_1 + \dfrac{d\sigmap}{\sqrt{\nproxy}}}$ & Lower \\ \addlinespace[0.5em] \midrule \addlinespace[0.5em]
Truncated Joint & ~~$\max\bc{\dfrac{s \sigmag }{\sqrt{\ngold}}\log\bp{d\ngold},~ \dfrac{sd\sigmap }{\sqrt{\nproxy}}\log\bp{d\nproxy}}$ & Upper \\ \addlinespace[0.5em] \bottomrule
\end{tabular}
\caption{Comparison of parameter estimation error across estimators.}
\label{tab:bounds}
\end{table}
\textit{Comparison:} For ease of comparison, we tabulate the bounds we have derived so far (up to constants and logarithmic factors) in Table \ref{tab:bounds}. Recall that we are interested in the regime where $\nproxy$ is large and $\ngold$ is small. Even with infinite proxy samples, the proxy estimator's error is bounded below by its bias $\bn{\delta^*}_1$. The gold estimator's error can also be very large, particularly when $\ngold \lesssim d$. Model averaging and weighted loss functions do not improve this picture by more than a constant factor. Now, note that in our regime of interest,
\[ \frac{s \sigmag }{\sqrt{\ngold}} ~\ll~ \frac{d \sigmag }{\sqrt{\ngold}} \text{~~~ and ~~~} \frac{sd \sigmap }{\sqrt{\nproxy}} ~\ll~ \bn{\delta^*}_1 + \frac{d \sigmap }{\sqrt{\nproxy}} \,.\] 
The first claim follows when $s\ll d$ (i.e., the bias term $\delta^*$ is reasonably sparse), and the second claim follows when $\bn{\delta^*}_1 \gg sd \sigmap/\sqrt{\nproxy}$ (i.e., the proxy estimator's error primarily arises from its bias $\delta^*$ rather than its variance, and $s$ is small). Thus, the joint estimator's error can be significantly lower than popular heuristics in our regime of interest.

\textit{Sample Complexity:} We can also interpret these results in terms of sample complexity. Let the decision-maker target a nontrivial parameter estimation error $\xi < \bn{\delta^*}_1$. As noted earlier, even an infinite number of proxy observations will not suffice, and one requires some gold data. Based on the bounds in Table \ref{tab:bounds}, it can easily be verified that the gold OLS/ridge, model averaging and weighted loss functions require $\ngold = \cO\bp{d^2 \sigmag^2 / \xi^2}$ \textit{regardless} of the number of proxy observations. In contrast, the joint estimator only requires $\ngold = \cO\bp{s^2\sigmag^2 \log^2\bp{d \cdot \frac{\sigmag}{\xi}}/\xi^2}$ as long as $\nproxy \gtrsim \mathcal{O}\bp{s^2 d^2\sigmap^2 \log^2\bp{d \cdot \frac{\sigmap}{\xi}}/\xi^2}$. In other words, when sufficient proxy data is available, the number of gold observations required is exponentially smaller in the dimension $d$. 

\subsection{Proof of Theorem \ref{thm:main}} \label{ssec:proof}

We start by defining the following two events:
\begin{equation} \label{eq:J}
\J = \bc{ \frac{2}{\ngold} \bn{ \epg^\top \Xg }_\infty \leq \lambda_0 } \,,
\end{equation}
\begin{equation} \label{eq:I}
\I = \bc{ \bn{\Xp^\top\epp}_2^2 \leq \lambda_1 } \,,
\end{equation}
where we have introduced two new parameters $\lambda_0$ and $\lambda_1$. We denote the complements of these events as $\J^C$ and $\I^C$ respectively. When events $\J$ and $\I$ hold, the gold and proxy noise terms $\epg$ and $\epp$ are bounded in magnitude, allowing us to bound our parameter estimation error $\bn{\tdelta - \hdelta}_1$. Since our noise is subgaussian, $\J$ and $\I$ hold with high probability (Lemmas \ref{lem:J} and \ref{lem:I}). We will choose the parameters $\lambda_0$ and $\lambda_1$ later to optimize our bounds.

\begin{lemma} \label{lem:basic-ineq}
On the event $\J$, taking $\lambda \geq 5 \lambda_0$, the solution $\hdelta$ to the optimization problem (\ref{eq:obj2}) satisfies
\[ \lambda \bn{ \tdelta - \hdelta }_1 ~\leq~ \frac{5}{4\ngold} \bn{ \Xg \nu }_2^2 + 5 \lambda \bn{ \nu }_1 + \frac{5\lambda^2 s}{2 \phi^2} \,. \]
\end{lemma}

\proof{Proof of Lemma \ref{lem:basic-ineq}}
Since the optimization problem (\ref{eq:obj2}) is convex, it recovers the in-sample global minimum. Thus, we must have that 
\[ \frac{1}{\ngold} \bn{\Yg - \Xg \bp{ \hdelta + \hbetap}}_2^2 + \lambda \bn{ \hdelta }_1 ~\leq~ \frac{1}{\ngold} \bn{\Yg - \Xg \bp{ \tdelta + \hbetap} }_2^2 + \lambda \bn{ \tdelta }_1 \,.\]
Substituting $\Yg = \Xg \betag + \epg = \Xg\bp{\tdelta +\hbetap} + \epg$ yields
\[ \frac{1}{\ngold} \bn{\Xg\bp{\tdelta -\hdelta} + \epg }^2_2 + \lambda\bn{\hdelta }_1 ~\leq~ \frac{1}{\ngold} \bn{\epg }^2_2 + \lambda\bn{\tdelta }_1 \,. \]
Expanding $\bn{\Xg \bp{\tdelta -\hdelta} + \epg }^2_2 = \bn{\Xg\bp{\tdelta - \hdelta} }^2_2 + \bn{ \epg }^2_2 + 2 \epg^\top \Xg \bp{\tdelta - \hdelta}$ and cancelling terms on both sides gives us
\begin{equation} \label{eq:basic-eq}
\frac{1}{\ngold} \bn{\Xg\bp{\tdelta - \hdelta} }^2_2 + \lambda \bn{ \hdelta }_1 ~\leq~ \frac{2}{\ngold}  \epg^\top \Xg \bp{\hdelta - \tdelta} + \lambda \bn{ \tdelta }_1 \,.
\end{equation}
By H\"older's inequality, when $\J$ holds and $\lambda \geq 5 \lambda_0$, we have
\begin{align*}
\frac{2}{\ngold}  \epg^\top \Xg \bp{\hdelta - \tdelta} ~&\leq~ \frac{2}{\ngold} \bn{ \epg^\top \Xg }_\infty \cdot \bn{ \hdelta - \tdelta }_1 \\
&\leq~ \frac{\lambda}{5} \bn{ \hdelta - \tdelta }_1 \,.
\end{align*}
Substituting into Eq. (\ref{eq:basic-eq}), we have on $\J$ that
\begin{align} \label{eq:basic-on-J}
\frac{5}{\ngold} \bn{\Xg\bp{\tdelta - \hdelta} }^2_2 + 5 \lambda \bn{ \hdelta }_1 ~&\leq~ \lambda \bn{\hdelta - \tdelta }_1 + 5\lambda \bn{ \tdelta }_1  \nonumber \\
&=~ \lambda \bn{\hdelta - \delta^* + \nu }_1 + 5\lambda \bn{ \delta^* - \nu }_1 \,,
\end{align}
where we recall that $\nu = \delta^* - \tdelta$. The second line uses $\nu$ to express the right hand side in terms of $\delta^*$ so that we can ultimately invoke the compatibility condition on $\hdelta$ (Definition \ref{def:cc}). To do this, we must first express $\hdelta$ in terms of its components on the index set $S = supp(\delta^*)$.

By the triangle inequality, we have
\begin{align}
\bn{ \hdelta }_1 ~&=~ \bn{\hdelta_S }_1 + \bn{ \hdelta_{S^c} }_1 \nonumber \\
~&\geq~ \bn{\delta_S^* }_1 - \bn{ \hdelta_S - \delta_S^* }_1 +  \bn{\hdelta_{S^c} }_1 \,. \label{eq:triangle-1}
\end{align}
Similarly, noting that $\delta_{S^c}^* = 0$ by definition of $S$, we have
\begin{align}
\bn{\hdelta - \delta^* + \nu }_1 ~&\leq~ \bn{ \hdelta_S - \delta_S^* }_1 + \bn{\hdelta_{S^c} }_1 + \bn{ \nu }_1 \,. \label{eq:triangle-2}
\end{align}
Collecting Eq. \eqref{eq:triangle-1}--\eqref{eq:triangle-2} and substituting into Eq. (\ref{eq:basic-on-J}), we have that when $\J$ holds,
\begin{equation} \label{eq:basic-on-J-2}
\frac{5}{\ngold} \bn{\Xg \bp{\tdelta - \hdelta} }^2_2 + 4 \lambda \bn{\hdelta_{S^c} }_1 ~\leq~ 6 \lambda \bn{\hdelta_S - \delta_S^* }_1 + 6\lambda \bn{ \nu }_1 \,.
\end{equation}

Ideally, we would now invoke the compatibility condition (Definition \ref{def:cc}) to $u = \hdelta - \delta^*$ to bound $\bn{\hdelta_S - \delta_S^* }_1$ in Eq. \eqref{eq:basic-on-J-2} above. However, this requires $u$ to satisfy $ \| u_{S^c} \|_1 \leq 3 \|u_S\|_1$, which may not hold in general. Thus, we proceed by considering two cases: either (i) $\bn{ \nu }_1 \leq \bn{\hdelta_S - \delta_S^* }_1$, or (ii) $\bn{\hdelta_S - \delta_S^* }_1 < \bn{ \nu }_1 $. In Case (i), we will invoke the compatibility condition to prove our finite-sample guarantee for the joint estimator, and in Case (ii), we will find that we already have good control over the error of the estimator.

\textbf{Case (i):} We are in the case that $\bn{ \nu }_1 \leq \bn{\hdelta_S - \delta_S^* }_1$, so from Eq. (\ref{eq:basic-on-J-2}), we can write on $\J$,
\[ \frac{5}{\ngold} \bn{\Xg \bp{\tdelta - \hdelta} }^2_2 + 4 \lambda \bn{\hdelta_{S^c} }_1 ~\leq~ 12 \lambda \bn{\hdelta_S - \delta_S^* }_1 \,. \]
Dropping the first (non-negative) term on the left hand side, we immediately observe that
\[ \bn{\hdelta_{S^c} }_1 = \bn{\hdelta_{S^c}  - \delta^*_{S^c}}_1 ~\leq~ 3 \bn{\hdelta_S - \delta_S^* }_1 \,,\]
so we can apply the compatibility condition to $u = \hdelta - \delta^*$. This yields
\[ \bn{\hdelta_S  - \delta^*_S}^2_1 ~\leq~ \frac{s}{\phi^2}  \bp{\hdelta  - \delta^*} \Sigmag \bp{\hdelta  - \delta^*} \,. \]
Taking the square-root, we get
\begin{equation} \label{eq:case1-int}
\bn{\hdelta_S  - \delta^*_S}_1 ~\leq~ \frac{\sqrt{s}}{\phi \sqrt{\ngold}} \bn{ \Xg \bp{\hdelta  - \delta^* } }_2 \,.
\end{equation}
Separately, when Case (i) and $\J$ hold, we can further simplify
\begin{align*}
\frac{5}{\ngold} \bn{ \Xg \bp{\tdelta - \hdelta} }_2^2 + 4 \lambda \bn{ \tdelta - \hdelta }_1 ~&=~ \frac{5}{\ngold} \bn{ \Xg \bp{\tdelta - \hdelta} }_2^2 + 4\lambda \bn{\hdelta - \delta^* + \nu }_1 \\
&\leq~ \frac{5}{\ngold} \bn{ \Xg \bp{\tdelta - \hdelta} }_2^2 + 4\lambda \bn{\hdelta_S - \delta^*_S}_1 + 4\lambda \bn{ \hdelta_{S^c} }_1 \\
&\hspace{0.2in}+ 4 \lambda \bn{\nu }_1 \\
&\leq~ 10 \lambda \bn{ \hdelta_S - \delta^* }_1 + 10 \lambda \bn{\nu }_1\,,
\end{align*}
where we used Eq. \eqref{eq:triangle-2} in the first inequality and Eq. (\ref{eq:basic-on-J-2}) in the second inequality. We can now proceed by applying Eq. (\ref{eq:case1-int})
\begin{align*}
\frac{5}{\ngold} \bn{ \Xg \bp{\tdelta - \hdelta} }_2^2 + 4 \lambda \bn{ \tdelta - \hdelta }_1 ~&\leq~ \frac{10 \lambda \sqrt{s}}{\phi \sqrt{\ngold}} \bn{ \Xg \bp{\hdelta  - \delta^* } }_2 + 10 \lambda \bn{ \nu }_1 \\
&\leq~ \frac{5}{2\ngold} \bn{ \Xg \bp{\hdelta - \delta^*} }_2^2 + 10 \lambda \bn{ \nu }_1 + \frac{10 \lambda^2 s}{ \phi^2} \\
&\leq~ \frac{5}{\ngold} \bn{ \Xg \bp{\tdelta - \hdelta} }_2^2 + \frac{5}{\ngold} \bn{ \Xg \nu }_2^2 + 10 \lambda \bn{ \nu }_1 \\
&\hspace{0.2in}+ \frac{10 \lambda^2 s}{\phi^2} \,,
\end{align*}
where the second and third inequalities follow from the facts that $10ab \leq 5a^2/2 + 10b^2$ and $(a+b)^2 \leq 2a^2 + 2b^2$ for any $a,b \in \R$. Then, when $\J$ and Case (i) hold, we have that
\begin{equation} \label{eq:case1}
\lambda \bn{ \tdelta - \hdelta }_1 ~\leq~ \frac{5}{4\ngold} \bn{ \Xg \nu }_2^2 + \frac{5\lambda}{2} \bn{ \nu }_1 + \frac{5 \lambda^2 s}{2 \phi^2} \,. 
\end{equation}

\textbf{Case (ii):} We are in the case that $\bn{\hdelta_S - \delta_S^* }_1 \leq \bn{ \nu }_1$, so Eq. (\ref{eq:basic-on-J-2}) implies on $\J$,
\begin{align}
\frac{5}{\ngold} \bn{\Xg \bp{\tdelta - \hdelta} }^2_2 + 4 \lambda \bn{\hdelta_{S^c} }_1 ~\leq~ 12 \lambda \bn{\nu }_1 \,. \label{eq:case2-int}
\end{align}
In this case, we do not actually need to invoke the compatibility condition. When $\J$ and Case (ii) hold, we can directly bound
\begin{align*} 
\frac{5}{\ngold} \bn{ \Xg \bp{\tdelta - \hdelta} }_2^2 + 4 \lambda \bn{ \tdelta - \hdelta }_1 ~&\leq~ \frac{5}{\ngold} \bn{ \Xg \bp{\tdelta - \hdelta} }_2^2 + 4\lambda \bn{\hdelta_S - \delta^*_S}_1 + 4\lambda \bn{ \hdelta_{S^c} }_1 \\
&\hspace{.2in} + 4 \lambda \bn{\nu }_1 \\
&\leq~ 20 \lambda \bn{ \nu }_1 \,,
\end{align*}
where we used Eq. \eqref{eq:triangle-2} in the first inequality and Eq. \eqref{eq:case2-int} as well as the fact that $\bn{\hdelta_S - \delta_S^* }_1 \leq \bn{ \nu }_1$ in the second inequality. Dropping the first (non-negative) term on the left hand side yields
\begin{align} \label{eq:case2}
\lambda \bn{ \tdelta - \hdelta }_1 ~&\leq~ 5 \lambda \bn{ \nu }_1 \,.
\end{align}
Combining the inequalities from Eq. (\ref{eq:case1}) and (\ref{eq:case2}), the following holds in both cases on $\J$,
\begin{align}
\label{eq:combine}
\lambda \bn{ \tdelta - \hdelta }_1 ~\leq~ \frac{5}{4\ngold} \bn{ \Xg \nu }_2^2 + 5 \lambda \bn{ \nu }_1 + \frac{5\lambda^2 s}{2 \phi^2} \,.
\end{align}
\Halmos
\endproof

In other words, we have shown that we can bound $\bn{ \tdelta - \hdelta }_1$ with high probability when $\nu$ (the approximation error of $\hbetap$) is small. We expect this error to be small since the number of proxy samples $\nproxy$ is large. The next lemma bounds the terms that depend on $\nu$ on the event $\I$. The proof is given in Appendix \ref{app:missing}.

\begin{lemma} \label{lem:bound-nu}
On the event $\I$, we have that both
\begin{align*}
\bn{\Xg\nu}_2^2 \le \frac{d\ngold}{\psi^2\nproxy^2}\lambda_1 \,,\quad\quad \text{and} \quad\quad
\bn{\nu}_1 \leq \frac{\sqrt{d\lambda_1}}{\psi \nproxy} \,.
\end{align*}
\end{lemma}

The next lemma simply applies these bounds on $\nu$ to the bound we derived earlier on $\bn{ \tdelta - \hdelta }_1$.

\begin{lemma} \label{lem:basic-ineq-2}
On the events $\J$ and $\I$, taking $\lambda \geq 5\lambda_0$, the solution $\hdelta$ to the optimization problem (\ref{eq:obj2}) satisfies
\[ \bn{ \tdelta - \hdelta }_1 ~\leq~ \frac{5d\lambda_1}{4\psi^2\nproxy^2\lambda} + \frac{5\sqrt{d\lambda_1}}{\psi \nproxy} + \frac{5\lambda s}{2 \phi^2} \,.\]
\end{lemma}

\proof{Proof of Lemma \ref{lem:basic-ineq-2}} 
The bound follows from applying Lemma \ref{lem:bound-nu} to the result in Lemma \ref{lem:basic-ineq}.
\Halmos
\endproof

Lemma \ref{lem:basic-ineq-2} shows that we can bound our parameter estimation error on the events $\J$ and $\I$. The next two lemmas use a concentration inequality for subgaussian random variables to show that these events hold with high probability. Their proofs are given in Appendix \ref{app:missing}.

\begin{lemma} \label{lem:J} The probability of event $\J$ is bounded by
\[ \Pr\bb{\J} ~\geq~ 1- 2d \exp\bp{-\frac{\lambda_0^2 \ngold}{8 \sigmag^2}} \,.\]
\end{lemma}

\begin{lemma} \label{lem:I} The probability of event $\I$ is bounded by
\[ \Pr\bb{\I} ~\geq~ 1- 2d \exp\bp{-\frac{\lambda_1}{2 d \sigmap^2 \nproxy}} \,.\]
\end{lemma}

We now combine Lemmas \ref{lem:basic-ineq-2}, \ref{lem:J}, and \ref{lem:I}, and choose values of our parameters $\lambda_0$ and $\lambda_1$ to complete our proof of Theorem \ref{thm:main}.

\proof{Proof of Theorem \ref{thm:main}} 
By Lemma \ref{lem:basic-ineq-2}, the following holds with probability $1$ when the events $\J$ and $\I$ hold, and $\lambda \geq 5\lambda_0$
\[ \bn{ \tdelta - \hdelta }_1 ~\leq~ \frac{5d\lambda_1}{4\psi^2\nproxy^2\lambda} + \frac{5\sqrt{d\lambda_1}}{\psi \nproxy} + \frac{5\lambda s}{ 2\phi^2} \,. \]
Recall that $\lambda_0, \lambda_1$ are theoretical quantities that we can choose freely to optimize our bound. In contrast, $\lambda$ is a fixed regularization parameter chosen by the decision-maker when training the estimator. Then, setting $\lambda_0 = \lambda/5$, we can write
\begin{align*}
\Pr\bb{\bn{ \tdelta - \hdelta }_1 ~\geq~ \frac{5d\lambda_1}{4\psi^2\nproxy^2\lambda} + \frac{5\sqrt{d\lambda_1}}{\psi \nproxy} + \frac{5\lambda s}{ 2\phi^2}} &~\leq~ 1 - \Pr[\J \cap \I] \\
&~\leq~ \Pr[\J^C] + \Pr[\I^C] \\
&~\leq~ 2d \exp\bp{-\frac{\lambda^2 \ngold}{200 \sigmag^2}} + 2d \exp\bp{-\frac{\lambda_1}{2 d \sigmap^2 \nproxy}} \,.
\end{align*}
The second inequality follows from a union bound, and the third follows from Lemma \ref{lem:J} (setting $\lambda_0 = \lambda/5$) and Lemma \ref{lem:I}.
By inspection, we choose
\[ \lambda_1 = \frac{\nproxy^2 \lambda^2}{d} \,,\]
yielding
\begin{align*}
\Pr\bb{\bn{ \tdelta - \hdelta }_1 ~\geq~ 5\lambda \bp{\frac{1}{4\psi^2} + \frac{1}{\psi} + \frac{s}{2 \phi^2} }} &~\leq~ 2d \exp\bp{-\frac{\lambda^2 \ngold}{200 \sigmag^2}} + 2d \exp\bp{-\frac{\lambda^2 \nproxy}{2 d^2 \sigmap^2}} \,.
\end{align*}
Finally, we reverse our variable transformation by substituting $\hbeta_{joint} = \hdelta + \hbetap$ and $\betag = \tdelta + \hbetap$, which gives us the result.
\Halmos
\endproof

\subsection{Nonlinear Predictors} \label{ssec:nonlinear}

Thus far, we have focused on linear predictors. Our joint estimator in Section \ref{ssec:joint-def} can be adapted to any $M$-estimator given its parametric empirical loss function $\ell(\cdot)$ as follows:
\begin{align}
\textbf{Step 1: ~~~}& \hbeta_{proxy} = \arg\min_{\beta}~ \bc{ \frac{1}{\nproxy} \sum_{i=1}^{\nproxy} \ell\bp{\beta;~\Xp^{(i)}, \Yp^{(i)}} } \,, \nonumber \\
\textbf{Step 2: ~~~}& \hbeta_{joint}(\lambda) = \arg\min_{\beta}~ \bc{ \frac{1}{\ngold} \sum_{i=1}^{\ngold} \ell\bp{\beta;~\Xg^{(i)}, \Yg^{(i)}} + \lambda\bn{\beta - \hbetap }_1 } \,. \label{eq:obj-glm}
\end{align}
Our proof techniques for bounding the resulting parameter estimation error generalize straightforwardly as long as $\ell(\cdot)$ is convex and satisfies mild technical assumptions \citep[in particular, the classical margin condition, e.g., see][]{negahban,buhlmann}. To illustrate, in this section, we extend our theoretical guarantees to the family of generalized linear models.

\textit{Preliminaries:} We start by defining some additional notation. Under the variable transformation $\delta = \beta - \hbetap$, let the in-sample loss function on the training set $(\mathbf{X}, \mathbf{Y})$ be
\[ \cL(\delta;~\mathbf{X}, \mathbf{Y}) = \frac{1}{n} \sum_{i=1}^{n} \ell\bp{\delta + \hbetap;~\mathbf{X}^{(i)}, \mathbf{Y}^{(i)}} \,.\]
Then, the second-stage objective \eqref{eq:obj-glm} can be re-written as
\begin{align}
\hdelta(\lambda) = \arg\min_\delta ~\cL(\delta;~\Xg, \Yg) + \lambda \bn{\delta}_1 \,. \label{eq:obj-glm2}
\end{align}
Once again, we define the optimal parameter $\tdelta = \min_\delta\E_{\epg}\cL(\delta)$ (recall that $\tdelta \neq \delta^*$ because we only have access to $\hbetap$ rather than $\betap$). We define the empirical process $w(\delta)$, and the expected error $\cE(\delta)$ relative to the expected error of the optimal parameter $\tdelta$ respectively as
\begin{align*}
w(\delta) &= \cL(\delta) - \E_{\epg} \cL(\delta) \,,\\
\cE(\delta) &= \E_{\epg}\cL(\delta) - \E_{\epg}\cL(\tdelta) \,.
\end{align*}
The margin condition essentially states that $\cE(\delta)$ is lower bounded by a positive quantity that scales with $\bn{\delta - \tdelta}$, i.e., the loss function $\ell(\cdot)$ evaluated on $(\Xg, \Yg)$ cannot be ``flat" around the optimal parameter $\tdelta$; if this is not the case, then we cannot distinguish the optimal parameter from nearby parameters on the training data. This is a standard assumption in the classification literature \citep{tsybakov2004optimal}, and has previously been adapted to studying nonlinear $M$-estimators in high dimension \citep{negahban}. Generalized linear models with strongly convex inverse link functions naturally satisfy a quadratic margin condition (see Lemma \ref{lem:E-bound}).

\textit{Generalized Linear Models:} A generalized linear model with parameter $\beta$ and observation $\vx$ has outcomes distributed as
\begin{align*}
y \sim \exp\bp{y \vx^\top \beta - A\bp{\vx^\top \beta} + B\bp{y}}\,,
\end{align*}
where $A$ and $B$ are known functions. Under this model, $\E[y \mid \vx] = \mu\bp{\vx^\top \beta}$ and $\text{Var}[y\mid \vx] = \mu'\bp{\vx^\top \beta}$, where $\mu = A'$ is the \textit{inverse link function}. For instance, in logistic regression, we have binary outcomes $y$ with $\mu(z) =1/(1 + \exp(-z))$; in Poisson regression, we have integer-valued outcomes $y$ with $\mu(z) = \exp(z)$; in linear regression, we have continuous outcomes $y$ with $\mu(z) = z$.

The resulting maximum likelihood estimator is $\hbeta = \arg\min_{\beta} \sum_{i=1}^n \bc{ -y_i \vx_i^\top \beta + A\bp{\vx_i^\top \beta} - B\bp{y_i }}$ \citep[see, e.g.,][]{mccullagh1989generalized}, implying the corresponding in-sample loss function
\begin{align} \label{eq:glm-ml}
\cL(\delta;~\mathbf{X}, \mathbf{Y}) &= \sum_{i=1}^n \bc{ -y_i \vx_i^\top \bp{\delta + \hbetap} + A\bp{\vx_i^\top \bp{\delta + \hbetap}} - B\bp{y_i} } \,.
\end{align}
In order to ensure convexity and the margin condition, we now impose that $A$ is strongly convex. This assumption is mild and is satisfied by any strictly increasing inverse link function, e.g., all the examples given above (logistic, Poisson, linear) satisfy strong convexity on a bounded domain. Note that Assumption \ref{ass:bounded} and our deterministic design matrix ensure that our domain is bounded.
\begin{assumption} \label{ass:glm}
The function $A(\cdot)$ is strongly convex with parameter $m > 0$, i.e., for all points $x,x'$ in its domain,
\[ A(x') - A(x) ~\geq~ A'(x) \cdot (x'-x) + m(x'-x)^2/2 \,. \]
\end{assumption}
As noted earlier, generalized linear models with strongly convex inverse link functions naturally satisfy a quadratic margin condition as follows.
\begin{lemma}[Margin Condition]\label{lem:E-bound}
For any $\delta$ in its domain, a generalized linear model satisfying Assumption \ref{ass:glm} also satisfies
\[ \cE(\delta) ~\geq~ \frac{m}{2\ngold} \bn{ \Xg \bp{\delta  - \tdelta } }_2^2 \,.\]
\end{lemma}
The proof of Lemma \ref{lem:E-bound} is given in Appendix \ref{app:glm-lemmas}. This result ensures that we can distinguish $\tdelta$ from other parameters on the training data when there is no noise.

We now have all the pieces required to establish a tail inequality that upper bounds the parameter estimation error of the two-step joint estimator with high probability for generalized linear models.
\begin{theorem} \label{thm:glm}
The nonlinear joint estimator satisfies the following tail inequality for a generalized linear model and any chosen value of the regularization parameter $\lambda > 0$:
\[ \Pr\bb{\bn{ \hbeta_{joint}(\lambda) - \betag  }_1 ~\geq~ \frac{5\lambda}{m} \bp{\frac{1}{8\psi^2} + \frac{1}{\psi} + \frac{s}{\phi^2} }} ~\leq~ 2d \exp\bp{-\frac{\lambda^2 \ngold}{50 \sigmag^2}} + 2d \exp\bp{-\frac{\lambda^2 \nproxy}{2 d^2 \sigmap^2}}  \,.\]
\end{theorem}

The proof of Theorem \ref{thm:glm} and associated lemmas is given in Appendix \ref{app:nonlinear}. Note that this result is nearly identical to the result in the linear case (Theorem \ref{thm:main}), with some small changes to the constants and an additional dependence on the strong convexity parameter $m$ related to the inverse link function. As a result, the expected parameter estimation error also satisfies the same bound as the linear case up to constants (see Corollary \ref{cor:glm} in Appendix \ref{app:glm-thm}).

\subsection{Remarks} \label{ssec:remarks}

We now briefly discuss applying our proposed estimator in practice.

\textit{Cross-validation:} While Corollary \ref{cor:joint} specifies a theoretically good choice for the regularization parameter $\lambda$, this choice depends on problem-specific parameters that are typically unknown. In practice, $\lambda$ is typically chosen using the popular heuristic of cross-validation \citep{picard1984cross, friedman2001}. In Appendix \ref{app:synth-exp}, we show that the two approaches attain very similar performance. A related literature investigates the consistency of estimators that use cross-validation for model selection \cite[see, e.g.,][]{li1987asymptotic, homrighausen2014leave}; future work could analogously study asymptotic properties of the joint estimator with cross-validation.

\textit{Scaling:} When the gold and proxy outcomes are different, it may be useful to perform a pre-processing step to ensure that both outcomes have similar magnitude. In Example \ref{ex:expedia}, clicks are roughly $10\times$ as frequent as purchases. Thus, in our numerical experiments, we scale down the responses $\Yp$ by this factor to ensure that the responses are of similar magnitude in expectation. i.e., $\E[|\yg(\vx)|] \approx \E[|\yp(\vx)|]$ for the same feature vector $\vx$. The scaling constant is typically known, or can be easily estimated from a hold-out set. While this step is not required for the theory, it helps increase the similarity between $\betag$ and $\betap$, making it more likely that we can successfully estimate the bias $\delta^*$ using a simple (sparse) function.

\textit{Combined estimation:} Our proposed two-step estimator does not require the simultaneous availability of gold and proxy data for training. This is an important feature in settings such as healthcare, where data from different sources often cannot be combined due to regulatory constraints. It also yields a simpler statistical analysis. However, if both sources of data are available together, one could alternatively combine the two-step estimation procedure, and directly estimate both $\hbetag(\lambda)$ and $\hbetap(\lambda)$ using the following heuristic:
\[ \small \bc{ \hbetag(\lambda),\hbetap(\lambda)} = \argmin_{\beta_{gold}, \beta_{proxy}}~ \bc{ \underbrace{\bn{\Yg - \Xg \beta_{gold} }^2_2}_{\Theta(\ngold)} +  \underbrace{\bn{\Yp - \Xp \beta_{proxy} }^2_2}_{\Theta(\nproxy)} + \lambda \bn{\beta_{proxy} - \beta_{gold} }_1 } \,. \]
We suggest choosing the regularization parameter $\lambda = \tO\bp{\sigmag \sqrt{\ngold} + \frac{d\sigmap\ngold}{\sqrt{\nproxy}}}$ to match the normalization suggested by Corollary \ref{cor:joint}. The expression above is for the linear case, but one can also consider nonlinear generalizations as in Section \ref{ssec:nonlinear}.

It is worth examining the regime where $\nproxy \gg \ngold$. In this case, the combined estimator actually decouples into our two-step procedure in Eq. \eqref{eq:obj}. This is because the second term, which scales as $\Theta(\nproxy)$, dominates the objective function so $\hbetap \approx \argmin_{\beta_{proxy}}\bn{\Yp - \Xp \beta_{proxy} }^2_2$. Once this value is fixed for $\hbetap$, the remaining optimization problem over $\beta_{gold}$ trivially reduces to the second step of our proposed estimator. Consistent with this observation, we simulate the combined estimator in Appendix \ref{app:synth-exp}, and find that its performance essentially matches that of the two-step joint estimator. Next, note that the first term of the combined estimator's objective scales as $\Theta(\ngold)$ while the third term scales as $\Theta(\sqrt{\ngold})$ when $\nproxy \gg \ngold$. Thus, when $\ngold$ is small, the regularization term will be large relative to the first term so $\hbetag$ will be strongly regularized towards $\hbetap$; however, when $\ngold$ becomes large, the first term will dominate the third term so we will eventually obtain the OLS estimate $\hbetag \approx \argmin_{\beta_{gold}}\bn{\Yg - \Xg \beta_{gold} }^2_2$.

\section{Experiments} \label{sec:experiments}

We now test the performance of our proposed joint estimator against benchmark estimators on both synthetic and real datasets.

\subsection{Synthetic} \label{ssec:synth}

We will consider two cases: (i) a sparse bias term $\delta^*$ (matching our assumptions and analysis), and (ii) a non-sparse bias term $\delta^*$ (i.e., $s=d$).

\textit{Data Generation:} We set $\nproxy = 1000$, $\ngold = 150$, $n_{test} = 1000$, $d = 100$, and fix our true parameter $\betag = \one \in \R^d$. We generate our proxy observations $\Xp \in \R^{\nproxy \times d}$ from a multivariate normal distribution with mean $\zero$ and a random covariance matrix generated as follows: (i) draw a random matrix in $\R^{d \times d}$ whose entries are uniform random samples from $[0,1]$, (ii) multiply the resulting matrix with its transpose to ensure that it is positive-definite, and (iii) normalize it with its trace. We take $\Xg  \in \R^{\ngold \times d}$ to simply be the first $\ngold$ rows of $\Xp$, and we additionally generate a test set $\mathbf{X}_{test} \in \R^{n_{test} \times d}$ in the same way we generate $\Xp$. Our data-generating process is the simple linear model, with $\Yp = \Xg \bp{\betag - \delta^*} + \epp$, $\Yg = \Xg \betag + \epg$, and test set responses $Y_{test} = \mathbf{X}_{test} \betag + \vep_{test}$. All noise terms are $\epp, \epg, \vep_{test} \sim \N(0, 1)$.

We study both sparse and non-sparse realizations of $\delta^*$. In the sparse case, $\delta^*$ is a randomly drawn binomial vector $0.1 \times B(d, 0.1)$, i.e., $s \ll d$. In the non-sparse case, $\delta^*$ is a randomly drawn gaussian vector $\N(\zero, 0.15 \times I_d)$, i.e., $s = d$. These parameters were chosen to keep the performance of the proxy and gold estimators relatively similar in both cases.

\textit{Estimators:} The gold and proxy estimators are simply OLS estimators on gold and proxy data respectively. The averaging, weighted, and joint estimators require a tuning parameter $\lambda$: for these, we split the gold observations randomly, taking 70\% to be the training set and the remaining 30\% to be the validation set. We then train models with different values of $\lambda$ on the training set, and use the mean squared error on the validation set to choose the best value of $\lambda$ for each estimator in the final model \citep{friedman2001}. Finally, we consider an ``Oracle" benchmark that has advance knowledge of the true (random) bias term $\delta^*$, and adjusts the proxy estimator accordingly.

\textit{Evaluation:}  Our evaluation metric is the average out-of-sample prediction error $\frac{1}{n_{test}} \bn{Y_{test} - \hat{Y}}_2^2$, where $\hat{Y}$ are the predictions of an estimator on the test set $\mathbf{X}_{test}$. We average our results over 100 trials, where we randomly draw all problem parameters in each iteration.

\begin{figure}[h]
  \begin{subfigure}[b]{0.5\textwidth}
    \includegraphics[width=\textwidth]{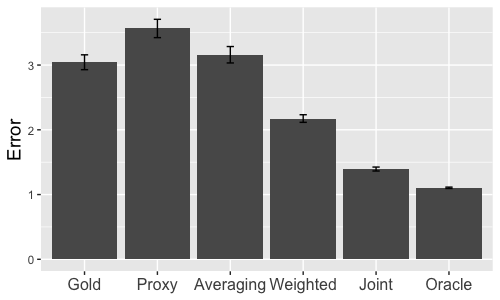}
    \caption{Sparse bias term}
    \label{fig:synth-1}
  \end{subfigure}
  \begin{subfigure}[b]{0.5\textwidth}
    \includegraphics[width=\textwidth]{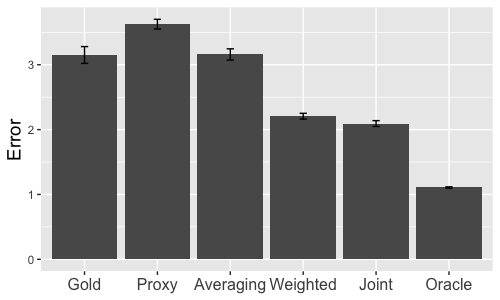}
    \caption{Non-sparse bias term}
    \label{fig:synth-2}
  \end{subfigure}
  \caption{Out-of-sample prediction error and 95\% confidence intervals of different estimators on synthetic data.}
\end{figure}

\textit{Results:} Figure \ref{fig:synth-1} shows results for the sparse bias term, while Figure \ref{fig:synth-2} shows results for the non-sparse bias term (error bars represent 95\% confidence intervals). We see that the joint estimator performs the best (excluding the oracle) in both the sparse and non-sparse cases. Sparsity does not appear to significantly affect the performance of any of the estimators except the joint estimator. In the sparse case, the joint estimator is very close to the oracle (i.e., we don't pay a significant price for not knowing the bias $\delta^*$), since we are able to recover a very close approximation of $\delta^*$. However, in the non-sparse case, the joint estimator yields only a slight improvement over the weighted estimator, and a sizeable gap remains between the oracle and the joint estimators (i.e., we pay a significant price for not knowing $\delta^*$). Thus, the joint estimator successfully leverages sparsity when present, but still performs comparably or better than popular heuristics otherwise.

\subsection{Recommendation Systems} \label{ssec:expedia}

Product variety has exploded, creating high search costs. As a consequence, many platforms offer data-driven recommendation systems that match customers with their preferred products. For example, Expedia is one of the world's largest online travel agencies, and serves millions of travelers a day. ``In this competitive market matching users to hotel inventory is very important since users easily jump from website to website. As such, having the best ranking of hotels for specific users with the best integration of price competitiveness gives an [online travel agency] the best chance of winning the sale" \citep{ICDM}. To inform these rankings, the goal is to train a model that can effectively predict which hotel rooms a customer will purchase.

In these settings, there are typically two outcomes: clicks and purchases. While purchases are the true outcome of interest, they are few and far between, making it hard to train an accurate model. On the other hand, clicks are much more frequent, and form a compelling proxy since customers will typically click on a product only if they have some intent to purchase. As a consequence, many recommendation systems use models that maximize click-through rates rather than purchase rates. In this case study, we take clicks and purchases as our proxy and true outcomes respectively.

\begin{figure}
\centering
\includegraphics[width=15cm]{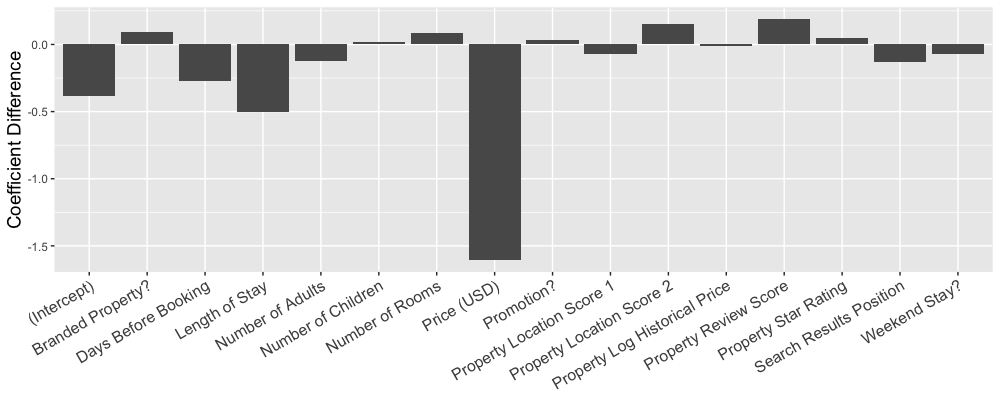}  
\caption{Difference in coefficients between a logistic regression predicting bookings and a logistic regression predicting clicks on the Expedia personalized recommendation dataset.}
\label{fig:expedia-coef-diff}
\end{figure}

\textit{Data:} We use personalized Expedia hotel search data that was made available through the 2013 International Conference on Data Mining challenge \citep{ICDM}. We only consider the subset of data where search results were randomly sorted to avoid position bias of Expedia's recommendation algorithm. After pre-processing, there are over 2.2 million customer impressions, 15 customer- and hotel-specific features related to the search destination, and 2 outcomes (clicks and bookings). We note that 0.05\% of impressions result in a click, while only 0.005\% result in a purchase. Thus, the gold outcomes are an order of magnitude more sparse than the proxy outcomes.

More details on data pre-processing and model training are given in Appendix \ref{app:expedia}.

\textit{Bias Term:} Since we have a very large number of observations, we can train accurate logistic regression models\footnote{We use logistic instead of linear regression since both outcomes are binary.} $\betap$ and $\betag$ for clicks and bookings respectively. Fig \ref{fig:expedia-coef-diff} shows the difference in the resulting parameter estimates, i.e., $\delta^* = \betag - \betap$. We immediately observe that the bias is in fact rather sparse --- nearly all the coefficients of $\delta^*$ are negligible (absolute value of the coefficient is relatively close to 0), with the notable exception of the price coefficient. Thus, our assumption that $\delta^*$ is sparse appears well-founded on this data. Moreover, we observe a systematic bias, where the hotel price negatively impacts bookings far more than clicks. Intuitively, a customer may not mind browsing expensive travel products, but is unlikely to make an expensive purchase. Thus, using predicted click-through rates alone (i.e., the proxy estimator) to make recommendations could result in overly expensive recommendations, thereby hurting purchase rates.

\begin{figure}[h]
\centering
\includegraphics[width=9cm]{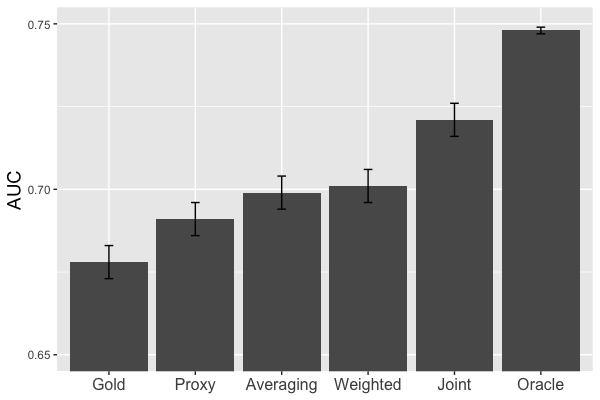}  
\caption{Predictive performance of different estimators on predicting bookings in the Expedia dataset.}
\label{fig:expedia-res}
\end{figure}

\textit{Setup:} In the data-rich regime (over 2 million impressions), the gold estimator is very accurate and there is no need for a proxy. We wish to study the data-scarce regime where proxies add value, so we restrict ourselves to small random subsamples of 10,000 impressions. Since we have binary outcomes, we use logistic rather than linear estimators. Note that $\ngold = \nproxy$, but the scarcity of bookings relative to clicks implies that $\sigmap \ll \sigmag$. Similar to the previous subsection, the averaging, weighted, and joint estimators are trained on a training set, and their tuning parameters are optimized over a validation set. Our oracle is the gold estimator trained on the full Expedia dataset (over 2 million impressions) rather than the small subsample. Since we no longer have access to the true parameter, we use predictive performance on a held-out test set to assess the performance of our different estimators. Performance is measured by AUC (area under ROC curve), which is more reliable than accuracy\footnote{Accuracy is a poor metric for imbalanced data: a trivial estimator that always predicts $0$ (no purchase) achieves 99.4\% accuracy, since 99.4\% of users do not purchase anything. A decision-maker would prefer an estimator that minimizes false positives while maintaining some baseline true positive rate (i.e., identifying most purchasing customers).} in imbalanced data \citep{friedman2001}. We average our results over 100 trials, where we randomly draw our training and test sets in each iteration.

\textit{Results:} Figure \ref{fig:expedia-res} shows the average performance on a held-out test set (error bars represent 95\% confidence intervals). We see that the joint estimator performs the best (excluding the oracle). In particular, it bridges half the gap between the best baseline (weighted estimator) and the oracle. In roughly 70\% of the trials, the joint estimator identifies price as a source of bias in $\hdelta$.

\subsection{Medical Risk Scoring} \label{ssec:diabetes}

A key component of healthcare delivery is patient risk scoring. Identifying patients who are at risk for a particular adverse event can help inform early interventions and resource allocation. In this case study, we consider Type II diabetes. In 2012, approximately 8.3\% of the world's adult population had diabetes, which is a leading cause of cardiovascular disease, renal disease, blindness, and limb amputation \citep{lall2017personalized}. To make matters worse, an estimated 40\% of diabetics in the US are undiagnosed, placing them at risk for major health complications \citep{cowie2009full}. At the same time, several clinical trials have demonstrated the potential to prevent type II diabetes among high-risk individuals through lifestyle interventions \citep{tuomilehto2011long}. Thus, our goal is to accurately predict patient-specific risk for Type II diabetes to inform interventions.

There are typically two ways a healthcare provider can obtain a risk predictor: train a new risk predictor based on its own patient cohort (true cohort of interest), or use an existing risk predictor that has been trained on a different patient cohort (proxy cohort) at a different healthcare provider. Training a new model can bring with it data scarcity challenges for small- or medium-sized providers; on the other hand, implementing an existing model can be problematic due to differences in physician behavior, shifts in patient characteristics, and discrepancies from how data is encoded in the medical record. In this case study, we will take patient data from a medium-sized provider as our gold data, and patient data pooled from two larger providers as our proxy data.

\textit{Data:} We use electronic medical record data across several healthcare providers. After basic pre-processing, we have roughly 100 features constructed from patient-specific information available \textit{before} his/her most recent visit, and our outcome is an indicator variable for whether the patient was diagnosed with diabetes \textit{during} his/her most recent visit. There are 980 patients in the proxy cohort (other providers), and 301 patients in the gold cohort (target provider), i.e., $\nproxy \gg \ngold$.

More details on data pre-processing and model training are given in Appendix \ref{app:emr}.

\textit{Setup:} Once again we have binary outcomes, so we use logistic rather than linear estimators. Similar to the previous subsections, the averaging, weighted, and joint estimators are trained on a training set, and their tuning parameters are optimized over a validation set. Since we no longer have access to the true parameter, we use predictive performance on a held-out test set to assess the performance of our different estimators. Performance is measured by AUC (area under ROC curve), which is more reliable than accuracy\footnote{Again, due to data imbalance, a trivial estimator that always predicts $0$ (low diabetes risk) achieves 87\% accuracy, since 87\% of patients will not be diagnosed with diabetes. A decision-maker would prefer an estimator that minimizes false positives while maintaining some baseline true positive rate (i.e., identifying most diabetic patients).} in imbalanced data \citep{friedman2001}. We average our results over 100 trials, where we randomly draw our training and test sets in each iteration.

\begin{figure}[h]
\centering
\includegraphics[width=9cm]{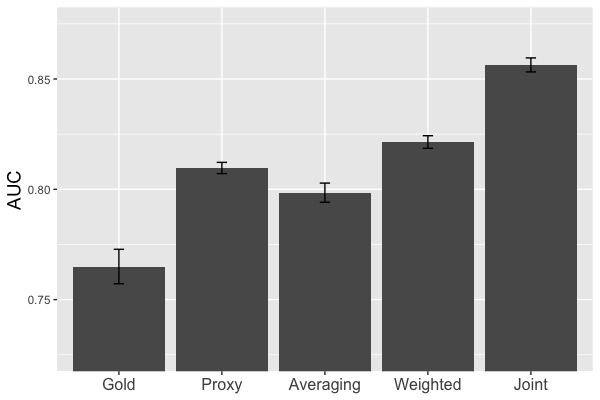}  
\caption{Predictive performance of different estimators on predicting diabetes in a medical record dataset.}
\label{fig:diabetes-res}
\end{figure}

\textit{Results:} Figure \ref{fig:diabetes-res} shows the average performance on a held-out test set (error bars represent 95\% confidence intervals). We see that the joint estimator performs the best by a significant margin.

\textit{Managerial Insights:} The improved performance of the joint estimator suggests that there is systematic bias as play between the proxy and gold patient cohorts. A better understanding of these biases can provide valuable managerial insights, and help inform better feature engineering and/or improved models for risk prediction. Accordingly, we study the estimated bias term $\hdelta$ across the 100 trials. We note that both the proxy and gold cohorts have similar rates of a diabetes diagnosis in the most recent visit: 14\% and 13\% respectively (the difference is not statistically significant). 

One feature that is frequently identified in $\hdelta$ is ICD-9 diagnosis code 790.21, which stands for ``Impaired fasting glucose." Impaired fasting glucose (also known as pre-diabetes) occurs when blood glucose levels in the body are elevated during periods of fasting, and is an important indicator for diabetes risk. Despite having similar diabetes diagnosis rates across the proxy and gold cohorts, 4.6\% of patients among the proxy cohort have an impaired fasting glucose diagnosis, while only 0.6\% of patients among the gold cohort have this diagnosis. Conversations with a physician suggest that physicians at the target healthcare provider (gold cohort) may not wish to burden the patients with fasting, which is required to diagnose a patient with impaired fasting glucose; in contrast, physicians at the proxy healthcare providers appear more willing to do so. As a consequence, ICD-9 code 790.21 is a highly predictive feature in $\betap$ for the proxy patient cohort, but not in $\betag$ for the gold patient cohort. Thus, differences in physician behavior can yield a systematic bias in the electronic medical records, and the joint estimator attempts to uncover such biases.

Similarly, another frequently identified feature is ICD-9 diagnosis code 278.0, which stands for ``Overweight and obesity." Again, despite having similar diabetes diagnosis rates across the proxy and gold cohorts, 5.6\% of patients among the proxy cohort have an obesity diagnosis, while only 0.9\% of patients among the gold cohort have this diagnosis. However, there is no significant difference in the recorded patient BMIs across proxy and gold cohorts, suggesting that the difference in obesity \textit{diagnosis} rates is not indicative of an actual difference in patient obesity rates. Conversations with a physician indicate that there are significant differences in how medical coders (staff responsible for encoding a patient's charts into the electronic medical record) choose which ICD-9 codes are recorded. As a consequence, ICD-9 code 278.0 is a highly predictive feature in $\betap$ for the proxy patient cohort, but not in $\betag$ for the gold patient cohort. Thus, differences in how patient chart data is encoded in the medical record can also yield systematic biases, which the joint estimator attempts to uncover.

Apart from these examples, the estimated bias $\hdelta$ also revealed differences in physician prescribing patterns. These biases are successfully leveraged by the joint estimator to improve performance.

\section{Discussion and Conclusions}

Proxies are copious and widely used in practice. However, the bias between the proxy predictive task and the true task of interest can lead to sub-optimal decisions. In this paper, we seek to \textit{transfer} knowledge from proxies to the true task by imposing sparse structure on the bias between the two tasks. We propose a two-step estimator that uses techniques from high-dimensional statistics to efficiently combine a large amount of proxy data and a small amount of true data. Our estimator provably achieves the same accuracy as popular heuristics with up to exponentially less gold data.

Proxy data is often viewed as a means of improving predictive accuracy. However, even with infinite proxy samples, the proxy estimator's error is bounded below by its bias $\bn{\delta^*}_1$. We propose that the true value of proxy data can actually lie in \textit{enhancing} the value of gold data. For instance, consider the case where $\ngold \lesssim \cO\bp{d^2 \sigmag^2}$. Our bounds show that the resulting gold OLS/ridge estimator's error is $\cO(1)$. In other words, without proxy data, limited gold data offers no predictive value. Often, additional gold data can be very costly or impossible to obtain, explaining the frequent reliance on alternative (proxy) data sources. However, when we have sufficient proxy data, i.e., $\nproxy \gtrsim \tO\bp{s^2 d^2 \sigmap^2}$, we only require $ \tO\bp{s^2 \sigmag^2 } \ll \cO\bp{d^2 \sigmag^2}$ gold observations to attain a nontrivial estimation error. Thus, proxy data can help us more efficiently use gold data: instead of using the limited gold data directly for estimating the predictive model, our estimator uses gold data to efficiently de-bias the proxy estimator. This insight can inform experimental design, particularly when decision-makers trade off the costs for obtaining labeled proxy and gold data.

Recovering the bias term also yields important managerial insights. For instance, it can be very difficult for hospital management to discover the systematic differences in physician diagnosing behavior or data recording across hospitals. As discussed in Section \ref{sec:experiments}, our estimator can recover an estimate of the bias term, which can shed light on the source of these biases. Once we understand these biases, one can perform better feature engineering, e.g., in the diabetes risk prediction example (Section \ref{ssec:diabetes}), we may learn to use the BMI feature instead of the obesity diagnosis feature. Knowing the bias between the proxies may also help us identify better sources of proxy data, e.g., in medical risk prediction, we may try to use patient data from a hospital with diagnosing patterns that are more similar to those in the target hospital.

There are a number of future research questions expanding the framework proffered in this paper. One important direction is to design transfer learning methods that can learn from a large number of (noisy) proxies, each with its own bias term. In line with our two case studies, such methods may be useful for online platforms that observe several proxies for a customer's intent to purchase (e.g., page views, clicks, and cart-adds), or for hospitals that have access to data from many peer hospitals. From a practical standpoint, it may also be interesting to extend these results to richer model families (e.g., neural networks or random forests) or exploring other types of (non-sparse) structure between the gold and proxy parameters (e.g., VC dimension or Rademacher complexity). Finally, a key prescriptive question involves studying the impact of transfer learning approaches on downstream decision-making in classic operations management problems. Some current work along these lines includes (i) dynamic pricing algorithms, where transfer learning from related products (proxies) are used to improve demand estimation (and therefore revenues) for the current product \citep{bastani2019meta}; and (ii) clinical trial designs, where transfer learning from early surrogate outcomes (proxies) are used to inform early stopping decisions in the trial \citep{anderer2019adaptive}.


{\SingleSpacedXI
\bibliographystyle{ormsv080} 
\bibliography{refs} 
}

%
%
%
\newpage
\begin{APPENDICES}

\section{Baseline Estimators} \label{app:baselines}

We now prove lower bounds on the parameter estimation error for various heuristics (Theorems \ref{thm:gold-ridge}--\ref{thm:weighted} from Section \ref{sec:baselines}). Since we are considering worst-case error over allowable $\bc{\Xg, \Xp, \betag, \betap, \epg, \epp}$, it suffices to consider a simple example where the assumptions and bounds hold. Here, we consider 
\begin{enumerate}
\item $\Xg$ is chosen such that $\Sigmag = \frac{1}{\ngold} \Xg^\top \Xg = I_d$ (where $I_d$ is the $d \times d$ identity matrix), and similarly, $\Xp$ is chosen such that $\Sigmap = \frac{1}{\nproxy} \Xp^\top \Xp = I_d$,
\item $\betag$ is chosen such that $\bn{\betag}_1 = 1$,
\item $\epg \sim \N(0,\sigmag^2 I_d)$ and $\epp \sim \N(0, \sigmap^2 I_d)$.
\end{enumerate}
Clearly, all assumptions made in Section \ref{sec:formulation} hold for this case.

\subsection{Proof of Theorem \ref{thm:gold-ridge}} \label{app:base-gold}

\proof{Proof of Theorem \ref{thm:gold-ridge}} We first show a lower bound for the OLS estimator, followed by the ridge estimator.

\textbf{(i) OLS Estimator:} The OLS estimator has the well-known closed form expression $\hbetag^{OLS} = \bp{\Xg^\top \Xg}^{-1}\Xg^\top \Yg$. Plugging in for $\Yg$ and $\Sigmag$ yields the (random) estimation error
\begin{align*}
\hbetag^{OLS} - \betag &= \bp{\Xg^\top \Xg}^{-1}\Xg^\top \epg \\
&= \frac{1}{\ngold} \Xg^\top \epg \,.
\end{align*}
Then, we can compute the variance
\begin{align*}
\var\bp{\hbetag^{OLS}-\betag} &= \frac{1}{\ngold^2} \E\bb{\Xg^\top \epg \epg^\top \Xg} \\
&= \frac{\sigma_{gold}^2}{\ngold} \,.
\end{align*}
Thus, using the distribution of $\epg$, we can write 
\[ \hbetag^{OLS}-\betag \sim \N\bp{0, \frac{\sigma_{gold}^2}{\ngold} I_d} \,. \]
Applying Lemma \ref{lem:l1-mu0-gaussian} in Appendix \ref{app:lemmas}, it follows that
\begin{align*}
\E\bb{\bn{\hbetag^{OLS} - \betag}_1} &= \tr\bp{I_d} \sqrt{\frac{2\sigma^2}{\pi \ngold}} \\
&= d \sqrt{\frac{2\sigma^2}{\pi \ngold}} \,.
\end{align*}
This computation gives us a lower bound of the parameter estimation error for the OLS estimator.

\textbf{(ii) Ridge Estimator:} Next, we consider the ridge estimator, which has the well-known closed form expression $\hbetag^{ridge}(\lambda) = \bp{\Xg^\top \Xg + \lambda I_d}^{-1}\Xg^\top \Yg$. Plugging in for $\Yg$ and $\Sigmag$ yields the (random) estimation error
\begin{align*}
\hbetag^{ridge}(\lambda) - \betag &= \bp{\Xg^\top \Xg + \lambda I_d}^{-1}\Xg^\top \Xg \betag + \bp{\Xg^\top \Xg + \lambda I_d}^{-1}\Xg^\top \epg - \betag  \\
&= \frac{\ngold}{\ngold + \lambda} \betag + \frac{1}{\ngold + \lambda} \Xg^\top \epg - \betag \\
&= -\frac{\lambda}{\ngold + \lambda} \betag + \frac{1}{\ngold + \lambda} \Xg^\top \epg \,.
\end{align*}
Then, we can compute the variance (note that the true parameter $\betag$ is not a random variable)
\begin{align*}
\var\bp{\hbetag^{ridge}(\lambda)-\betag} &= \frac{1}{(\ngold + \lambda)^2} \E\bb{\Xg^\top \epg \epg^\top \Xg} \\
&= \frac{\ngold \sigma_{gold}^2}{(\ngold + \lambda)^2} \,.
\end{align*}
Thus, using the distribution of $\epg$, we can write 
\[ \hbetag^{ridge}(\lambda)-\betag \sim \N\bp{-\frac{\lambda}{\ngold + \lambda} \betag, \frac{\ngold \sigma_{gold}^2}{(\ngold + \lambda)^2} I_d} \,. \]
Applying Lemma \ref{lem:l1-mu-gaussian} in Appendix \ref{app:lemmas}, it follows that
\begin{align*}
\E\bb{\bn{\hbetag^{ridge}(\lambda) - \betag}_1} &\geq \max \bc{ \frac{1}{2} \cdot \frac{\lambda}{\ngold + \lambda} \bn{\betag}_1, \sqrt{\frac{\ngold \sigma_{gold}^2}{2 \pi (\ngold + \lambda)^2}} \tr\bp{I_d}} \\
&= \max \bc{ \frac{1}{2} \cdot \frac{\lambda b}{\ngold + \lambda}, d \sqrt{\frac{\ngold \sigma_{gold}^2}{2 \pi (\ngold + \lambda)^2}} } \,.
\end{align*}
Note that the first term in the maximum is monotone increasing in $\lambda$, while the second term in the maximum is monotone decreasing in $\lambda$. Thus, the minimum value of the maximum is achieved when the two terms are equal, i.e., when
\[ \lambda = \frac{d}{b}\sqrt{\frac{2 \ngold \sigma_{gold}^2}{\pi}} \,.\]
Plugging in, we get
\begin{align*}
\min_{\lambda}~\E\bb{\bn{\hbetag^{ridge}(\lambda) - \betag}_1} &\geq \frac{ d\sigmag /\sqrt{2\pi}}{b\sqrt{\ngold} + d\sigmag \sqrt{2/\pi}} \,.
\end{align*}
Along with the previous case, this completes the proof.
\Halmos
\endproof

\subsection{Proof of Theorem \ref{thm:proxy-ols}} \label{app:base-proxy}

\proof{Proof of Theorem \ref{thm:proxy-ols}}
Note that the OLS estimator is $\hbetap = \bp{\Xp^\top \Xp}^{-1}\Xp^\top \Yp$. Plugging in for $\Yp$ and $\Sigmap$ yields the (random) estimation error
\begin{align*}
\hbetap - \betap &= \bp{\Xp^\top \Xp}^{-1}\Xp^\top \epp \\
&= \frac{1}{\nproxy} \Xp^\top \epp \,.
\end{align*}
Then, we can compute the variance
\begin{align*}
\var\bp{\hbetap-\betap} &= \frac{1}{\nproxy^2} \E\bb{\Xp^\top \epp \epp^\top \Xp} \\
&= \frac{\sigmap^2}{\nproxy} \,.
\end{align*}
Thus, using the distribution of $\epg$ and the fact that $\betag = \betap + \delta^*$, we can write 
\[ \hbetap-\betag \sim \N\bp{-\delta^*, \frac{\sigmap^2}{\nproxy} I_d} \,. \]
Applying Lemma \ref{lem:l1-mu-gaussian} in Appendix \ref{app:lemmas}, it follows that
\begin{align*}
\E\bb{\bn{\hbetag^{OLS} - \betag}_1} &\geq \max \bc{ \frac{1}{2} \bn{\delta^*}_1, \sqrt{\frac{\sigmap^2}{2\pi \nproxy}} \tr\bp{I_d}} \\
&= \max \bc{ \frac{1}{2} \bn{\delta^*}_1, d\sqrt{\frac{\sigmap^2}{2\pi \nproxy}}} \,.
\end{align*}
This computation gives us a lower bound of the parameter estimation error for the OLS estimator.
\Halmos
\endproof

\subsection{Proof of Theorem \ref{thm:avg}} \label{app:base-avg}
\proof{Proof of Theorem \ref{thm:avg}}
The averaging estimator can be expanded as
\begin{align*}
\hbeta_{avg}(\lambda) &= (1-\lambda) \cdot \hbetag^{OLS} + \lambda \cdot \hbetap \\
&= (1-\lambda) \cdot \bp{\Xg^\top \Xg}^{-1} \Xg^\top \Yg + \lambda \cdot \bp{\Xp^\top \Xp}^{-1} \Xp^\top \Yp \\
&= \frac{1-\lambda}{\ngold} \cdot \Xg^\top \Yg + \frac{\lambda}{\nproxy} \cdot \Xp^\top \Yp \\
&=  \frac{1-\lambda}{\ngold} \cdot \Xg^\top \bp{\Xg \betag + \epg} + \frac{\lambda}{\nproxy} \cdot \Xp^\top \bp{\Xp \betap + \epp}  \\
&= (1-\lambda) \betag + \lambda \betap + \frac{1-\lambda}{\ngold} \cdot \Xg^\top \epg + \frac{\lambda}{\nproxy} \cdot \Xp^\top \epp \,.
\end{align*}
Then,
\[ \hbeta_{avg}(\lambda) - \betag = \lambda \bp{\betap - \betag } + \frac{1-\lambda}{\ngold} \cdot \Xg^\top \epg + \frac{\lambda}{\nproxy} \cdot \Xp^\top \epp \,.\]
Using the fact that $\epg$ and $\epp$ are independent random variables, we can compute
\begin{align*}
\var\bp{\hbeta_{avg}(\lambda) - \betag} &= \bp{ \frac{1-\lambda}{\ngold} }^2 \var\bp{ \Xg^\top \epg} + \bp{\frac{\lambda}{\nproxy}}^2 \var\bp{ \Xp^\top \epp} \\
&=  \bp{\frac{(1-\lambda)^2 \sigmag^2}{\ngold} + \frac{\lambda^2 \sigmap^2}{\nproxy}} I_d \,.
\end{align*}
Thus, we have that
\[ \hbeta_{avg}(\lambda) - \betag \sim \N\bp{\lambda \bp{\betap - \betag }, \bp{\frac{(1-\lambda)^2 \sigmag^2}{\ngold} + \frac{\lambda^2 \sigmap^2}{\nproxy}} I_d} \,.\]
Applying Lemma \ref{lem:l1-mu-gaussian} in Appendix \ref{app:lemmas}, it follows that
\begin{align*}
\E\bb{\bn{\hbeta_{avg} - \betag}_1} &\geq \max \bc{ \frac{\lambda}{2} \bn{\betap - \betag}_1, \sqrt{\frac{(1-\lambda)^2 \sigmag^2}{2\pi \ngold} + \frac{\lambda^2 \sigmap^2}{2\pi \nproxy}} \tr\bp{I_d}} \\
&= \max \bc{ \frac{\lambda}{2} \bn{\delta^*}_1, d \sqrt{\frac{(1-\lambda)^2 \sigmag^2}{2\pi \ngold} + \frac{\lambda^2 \sigmap^2}{2\pi \nproxy}} } \\
&\geq \max \bc{ \frac{\lambda}{2} \bn{\delta^*}_1, d \sqrt{\frac{(1-\lambda)^2 \sigmag^2}{2\pi \ngold}}, d \sqrt{ \frac{\lambda^2 \sigmap^2}{2\pi \nproxy}} } \\
&\geq \frac{\lambda}{6} \bn{\delta^*}_1 + \frac{d}{3}\sqrt{ \frac{\lambda^2 \sigmap^2}{2\pi \nproxy}} + \frac{d}{3} \sqrt{\frac{(1-\lambda)^2 \sigmag^2}{2\pi \ngold}}  \,,
\end{align*}
where we have used the identity $\max\bc{a,b} \geq \frac{a+b}{2}$. Now, note that this expression is linear in $\lambda$, and thus, the value of $\lambda$ that minimizes the maximum occurs at one of the two extrema, i.e., $\lambda =0$ or $\lambda=1$. Then, we can write
\begin{align*}
\min_{\lambda}~\E\bb{\bn{\hbeta_{avg} - \betag}_1} &\geq \min\bc{ \frac{d \sigmag}{3\sqrt{2\pi \ngold}}, \frac{1}{6} \bn{\delta^*}_1 + \frac{d \sigmap}{3\sqrt{ 2\pi \nproxy}}  } \,.
\end{align*}
\Halmos
\endproof

\subsection{Proof of Theorem \ref{thm:weighted}} \label{app:base-wt}
\proof{Proof of Theorem \ref{thm:weighted}}
Recall that the weighted estimator is given by
\begin{align*}
\hbeta_{weight}(\lambda) &= \arg\min_{\beta} \bc{ \frac{1}{\lambda \ngold + \nproxy} \cdot \bp{ \lambda \| \Yg - \Xg \beta \|_2^2 + \| \Yp - \Xp \beta \|_2^2 }},.
\end{align*}
Setting the gradient to $0$, we get that $\hbeta_{weight}(\lambda)$ is the solution to the equation
\begin{align*}
\frac{2\lambda}{\lambda \ngold + \nproxy} \cdot \Xg^\top\bp{ \Yg - \Xg \beta } + \frac{2}{\lambda \ngold + \nproxy} \cdot \Xp^\top \bp{ \Yp - \Xp \beta } &= 0 \,.
\end{align*}
It is useful to define the variable 
\[\lambda' = \frac{\nproxy}{\lambda \ngold + \nproxy} \,,\]
and observe that
\[1-\lambda' = \frac{\lambda \ngold}{\lambda \ngold + \nproxy} \,.\]
Note that the allowed range of $\lambda \in [0,\infty)$ corresponds to the allowed range $\lambda' \in [0,1]$. Thus, we can equivalently write $\hbeta_{weight}$ as a function of $\lambda' \in [0,1]$ is the solution to
\begin{align*}
 \frac{1-\lambda'}{\ngold} \cdot \Xg^\top\bp{ \Yg - \Xg \beta } + \frac{\lambda'}{\nproxy} \cdot \Xp^\top \bp{ \Yp - \Xp \beta } &= 0 \,,
\end{align*}
which yields the solution
\begin{align*}
\hbeta_{weight}(\lambda') &= \bp{ \frac{1-\lambda'}{\ngold} \cdot \Xg^\top \Xg + \frac{\lambda'}{\nproxy} \cdot \Xp^\top \Xp}^{-1}\bp{\frac{1-\lambda'}{\ngold} \cdot \Xg^\top \Yg + \frac{\lambda'}{\nproxy} \cdot \Xp^\top \Yp} \\
&= \frac{1-\lambda'}{\ngold} \cdot \Xg^\top \Yg + \frac{\lambda'}{\nproxy} \cdot \Xp^\top \Yp \,.
\end{align*}
Note that this expression is exactly the averaging estimator (Section \ref{ssec:avg}) in this setting, and thus the proof and lower bound of Theorem \ref{thm:avg} apply directly. This completes the proof.
\Halmos
\endproof

\section{Linear Joint Estimator} \label{app:joint-linear}

\subsection{Missing Lemmas in Proof of Theorem \ref{thm:main}} \label{app:missing}

\proof{Proof of Lemma \ref{lem:bound-nu}}
Recall that 
\[ \nu ~=~ \hbetap-\betap ~=~ \bp{\Xp^\top \Xp}^{-1} \Xp^\top \epp \,.\]
Then, on event $\I$, we can write
\begin{align*}
\bn{\Xg\nu}_2^2 &= \bn{\Xg(\Xp^\top\Xp)^{-1}\Xp^\top\epp}_2^2 \\
&\leq \bn{\Xg}_2^2\cdot\bn{(\Xp^\top\Xp)^{-1}\Xp^\top\epp}_2^2 \\
&\le \frac{1}{\psi^2\nproxy^2}\bn{\Xg}_2^2\cdot\bn{\Xp^\top\epp}_2^2 \\
&\le \frac{d\ngold}{\psi^2\nproxy^2}\bn{\Xp^\top\epp}_2^2 \\
&\le \frac{d\ngold}{\psi^2\nproxy^2}\lambda_1 \,,
\end{align*}
where the first inequality follows from the definition of the matrix norm, the second inequality follows from Assumption \ref{ass:proxy-posdef} on the minimum eigenvalue of $\Sigmap$ yielding $\bp{\Xp^\top \Xp}^{-1} = \frac{1}{\nproxy}\Sigmap^{-1} \preceq \frac{1}{\psi \nproxy} I_d$, and the third inequality follows from the matrix norm identity that 
\[ \bn{\Xg}_2^2 ~\leq~ \tr\bp{\Xg^\top \Xg} ~=~ \ngold\tr\bp{\Sigmag} ~=~ d\ngold \,, \]
using the fact that we normalized $\text{diag}\bp{\Sigmag} = 1_{d\times 1}$. This proves the first inequality.

For the second inequality, we observe that on event $\I$,
\begin{align*}
\bn{\nu}_1 &= \bn{(\Xp^\top\Xp)^{-1}\Xp^\top\epp}_1 \\
&\leq \sqrt{d }\bn{(\Xp^\top\Xp)^{-1} \Xp^\top\epp}_2 \\
&\leq \sqrt{d }\bn{(\Xp^\top\Xp)^{-1} }_2 \cdot \bn{ \Xp^\top\epp}_2 \\
&\leq \frac{\sqrt{d}}{\psi \nproxy} \bn{ \Xp^\top\epp}_2 \\
&\leq \frac{\sqrt{d\lambda_1}}{\psi \nproxy} \,,
\end{align*}
where the first inequality follows from the definition of the matrix norm, the second inequality follows from Cauchy Schwarz, and the third inequality follows from Assumption \ref{ass:proxy-posdef}.
\Halmos
\endproof

\proof{Proof of Lemma \ref{lem:J}}
Recall from Eq. (\ref{eq:J}) that $\J = \bc{\dfrac{2}{\ngold}\bn{\epg^\top \Xg}_\infty \leq \lambda_0}$.  Then,
\begin{align*}
\Pr\bb{\J} &= 1-\Pr\bb{\max_{r \in [d]}~ \left|\epg^\top\Xg^{(r)} \right| ~\geq~ \frac{\lambda_0 \ngold}{2} } \\
&\geq 1 - d \cdot \Pr\bb{\left|\epg^\top\Xg^{(1)} \right| ~\geq~ \frac{\lambda_0 \ngold}{2} } \,,
\end{align*}
where $\Xg^{(r)} \in \R^d$ is the $r^{th}$ column of $\Xg$, and the first inequality follows from a union bound.

We can then apply Lemma \ref{lem:subg-sum} in Appendix \ref{app:lemmas} with $z_i = \bp{\epg}_i$, $a_i = \bp{\Xg}_{i,r}$, noting that our standardization of $\Xg$ implies each column satisfies 
\[ A = \sum_{i=1}^{\ngold} \bp{\Xg}_{i,r}^2 = \bn{\Xg^{(r)}}_2^2 = \ngold \,.\]
Then, by Lemma \ref{lem:subg-sum},
\[ W = \sum_{i=1}^{\ngold} \bp{\Xg}_{i,r} \bp{\epg}_i = \epg^\top \Xg^{(r)} \,,\]
is a $\bp{\sigmag\sqrt{\ngold}}$-subgaussian random variable. Thus, we can apply a concentration inequality for subgaussian random variables (Lemma \ref{lem:subg-conc} in Appendix \ref{app:lemmas}) to bound
\begin{align*}
\Pr\bb{\J} &\geq 1 - d \cdot \Pr\bb{\left|\epg^\top\Xg^{(1)} \right| ~\geq~ \frac{\lambda_0 \ngold}{2} } \\
&\geq 1-2d \exp\bp{-\frac{\lambda_0^2 \ngold}{8 \sigmag^2}} \,.
\end{align*}
\Halmos
\endproof

\proof{Proof of Lemma \ref{lem:I}} Recall from Eq. (\ref{eq:I}) that $\I = \bc{ \bn{\Xp^\top\epp}_2^2 \leq \lambda_1 }$. Then,
\begin{align*}
\bn{\Xp^\top\epp}_2^2 &= \bn{\sum_{i=1}^{\nproxy} \bp{\Xp}_i \bp{\epp}_i}_2^2 \\
&= \sum_{j=1}^d \bp{\sum_{i=1}^{\nproxy} \bp{\Xp}_{i,j} \epp^{(i)}}^2 \,,
\end{align*}
where $\bp{\Xp}_i \in \R^d$ is the $i^{th}$ row of $\Xp$, and $\bp{\epp}_i \in \R$ is the $i^{th}$ component of $\epp$. Note that $\bc{\epp^{(i)}}_{i=1}^{\nproxy}$ are independent $\sigmap$-subgaussian random variables. Thus, each summand is a linear combination of independent subgaussian random variables, which yields a new subgaussian random variable. For each $j \in [d]$, it is useful to define an intermediate variable 
\begin{align*}
\varepsilon_j'=\sum_{i=1}^{\nproxy} \bp{\Xp^{(i)}}_j \epp^{(i)}.
\end{align*}
Note that $\Xp^\top\epp \in \R^{d \times 1}$ is a vector whose elements are $\vep_j'$. We can now apply Lemma \ref{lem:subg-sum} in Appendix \ref{app:lemmas}, taking $\bc{z_i}$ to be $\bc{\epp^{(i)}}$, $\bc{a_i}$ to be $\bc{\bp{\Xp^{(i)}}_j}$, and noting that
\begin{align*}
A=\sum_{i=1}^{\nproxy} \bp{\Xp^{(i)}}_j\bp{\Xp^{(i)}}_j ~=~ \nproxy \Sigmap^{(jj)} ~=~ \nproxy\,,
\end{align*}
for each $j \in [d]$ since we have normalized $\text{diag}\bp{\Sigmap} = 1_{d\times 1}$. Then, by Lemma \ref{lem:subg-sum}, $\varepsilon_j'$ is $\bp{\sigmap \sqrt{\nproxy}}$-subgaussian. We can then apply a concentration inequality for subgaussian random variables (Lemma \ref{lem:subg-conc} in Appendix \ref{app:lemmas}) to bound
\begin{align*}
\Pr[\I] &= 1- \Pr \bb{\bn{\Xp^\top\epp}_2^2 ~\geq~ \lambda_1 } \\
&= 1-\Pr\bb{\bn{\vep'}_2^2 ~\geq~ \lambda_1} \\
&\geq 1- d \cdot \Pr\bb{|\vep'_j| ~\geq~ \sqrt{\frac{\lambda_1}{d}}} \\
&\geq 1-2d \exp\bp{-\frac{\lambda_1}{2 d \sigmap^2 \nproxy}} \,.
\end{align*}
\Halmos
\endproof

\subsection{Proof of Corollary \ref{cor:joint}} \label{app:corollary}

\proof{Proof of Corollary \ref{cor:joint}} For ease of notation, let 
\[ w = \frac{5}{4\psi^2} + \frac{5}{\psi} + \frac{5s}{2 \phi^2}  \,.\]
Recall from Lemma \ref{lem:basic-ineq-2} that
\[ \bn{ \hbeta_{joint}(\lambda) - \betag  }_1 ~\leq~  \lambda w\,,\]
with probability $1$ when the events $\J$ and $\I$ hold, and we take $\lambda_0 = \lambda/5$ and $\lambda_1 = \nproxy^2 \lambda^2/d$.

We expand the expected parameter estimation error
\begin{align} \label{eq:cor-exp}
\E\bb{\bn{\hbeta_{joint}^{tr}(\lambda) - \betag}_1} &= \E\bb{\bn{\hbeta_{joint}^{tr}(\lambda) - \betag}_1 ~\Big|~ \J \cap \I} \cdot \Pr[\J \cap \I] \nonumber \\
&~~~~+ \E\bb{\bn{\hbeta_{joint}^{tr}(\lambda) - \betag}_1 ~\Big|~ \J^C \cup \I^C} \cdot \Pr[\J^C \cup \I^C] \,.
\end{align}
To bound the first expectation on the right hand side of (\ref{eq:cor-exp}), we define a new event
\begin{align*}
B = \bp{\bn{\hbeta_{joint}(\lambda)}_1 \le 2b}.
\end{align*}
By definition, $\hbeta_{joint}^{tr} = \hbeta_{joint}$ when $B$ holds, and $\hbeta_{joint}^{tr} = 0$ otherwise. Then,
\begin{align*}
&\E\bb{\bn{\hbeta_{joint}^{tr}(\lambda) - \betag}_1 ~\Big|~ \J \cap \I} \\
&=~ \E\bb{\bn{\hbeta_{joint}^{tr}(\lambda) - \betag}_1 ~\Big|~ B \cap \J \cap \I} \cdot \Pr[B] + \E\bb{\bn{\hbeta_{joint}^{tr}(\lambda) - \betag}_1 ~\Big|~ B^C \cap \J \cap \I} \cdot \Pr[B^C] \\
&=~ \E\bb{\bn{\hbeta_{joint}(\lambda) - \betag}_1 ~\Big|~ B \cap \J \cap \I} \cdot \Pr[B] + \E\bb{\bn{\betag}_1 ~\Big|~ B^C \cap \J \cap \I} \cdot \Pr[B^C] \\
&\leq~  \lambda w \cdot\Pr[B] + \E\bb{\bn{\betag}_1 ~\big|~ B^C \cap \J \cap \I}\cdot\Pr[B^C] \,.
\end{align*}
Now, note that on the event $\bp{B^C \cap \J \cap \I}$, we have both that
\begin{align*}
&\bn{ \hbeta_{joint}(\lambda) - \betag  }_1 ~\leq~  \lambda w\,, \\
&\bn{\hbeta_{joint}(\lambda)}_1 ~\geq~ 2b ~\geq~ 2\bn{\betag}_1 \,.
\end{align*}
Together, these facts imply that on the event $\bp{B^C \cap \J \cap \I}$,
\begin{align*}
\bn{\betag}_1 &\leq \bn{\hbeta_{joint}(\lambda)}_1 - \bn{\betag}_1 \\
&\leq \bn{\hbeta_{joint}(\lambda) - \betag}_1 \\
&\leq \lambda w \,,
\end{align*}
using the triangle inequality. Thus,
\begin{align} \label{eq:cor-first}
\E\bb{\bn{\hbeta_{joint}^{tr}(\lambda) - \betag}_1 ~\Big|~ \J \cap \I} &\le \lambda w \cdot\Pr[B] + \E\bb{\bn{\betag}_1 ~\big|~ B^C \cap \J \cap \I}\cdot\Pr[B^C] \nonumber \\
&\leq \lambda w \cdot\Pr[B] + \lambda w \cdot\Pr[B^C]  \nonumber \\
&= \lambda w \,.
\end{align}
Next, we consider the second expectation on the right hand side of (\ref{eq:cor-exp}). Regardless of the events $\J, \I,$ and $B$, we always have the following bound
\begin{align} \label{eq:cor-second}
\bn{ \hbeta_{joint}^{tr}(\lambda) - \betag  }_1 ~&\leq~ \bn{ \hbeta_{joint}^{tr}(\lambda)}_1 + \bn{\betag  }_1 ~\leq~ 3b \,,
\end{align}
using the triangle inequality and the definition of $\hbeta_{joint}^{tr}(\lambda)$. Substituting (\ref{eq:cor-first}) and (\ref{eq:cor-second}) into (\ref{eq:cor-exp}), we have
\begin{align*}
\E\bb{\bn{\hbeta_{joint}^{tr}(\lambda) - \betag}_1} ~&\leq~ \lambda w \cdot \Pr[\J \cap \I] + 3b \cdot \Pr[\J^C \cup \I^C] \\
&\leq \lambda w + 3b \cdot \bp{\Pr[J^C] + \Pr[\I^C] } \\
&\leq \lambda w + 6bd \cdot \bp{\exp\bp{-\frac{\lambda^2 \ngold}{200 \sigmag^2}} + \exp\bp{-\frac{\lambda^2 \nproxy}{2 d^2 \sigmap^2}}}\,,
\end{align*}
using a union bound, and applying Lemmas \ref{lem:J} and \ref{lem:I} with the chosen values $\lambda_0 = \lambda/5$ and $\lambda_1 = \nproxy^2 \lambda^2/d$.
\Halmos
\endproof

\section{Nonlinear Joint Estimator} \label{app:nonlinear}

Throughout this section, we use the same notation as we did in the linear case. By definition, 
\[\betag ~=~ \hbetap + \min_\delta\E_{\epg}\cL(\delta) ~=~ \hbetap + \tdelta\,.\]
Thus, defining $\nu = \hbetap - \betap$ as before, we have $\tdelta = \delta^* -\nu$.

\subsection{Key Lemmas for GLMs} \label{app:glm-lemmas}

\proof{Proof of Lemma \ref{lem:E-bound}}
Using the definition of the empirical log likelihood $\cL$, we can expand
\begin{align*}
\cE(\delta) &= \E_{\epg} \bb{\cL(\delta) - \cL(\tdelta)} \\
&= \frac{1}{\ngold} \sum_{i=1}^{\ngold} \left\{-A'\bp{ \bp{\betag}^\top \Xg^{(i)}} \bp{\delta - \tdelta}^\top \Xg^{(i)} + A\bp{ \bp{\delta + \hbetap}^\top \Xg^{(i)}} - A\bp{ \bp{\betag}^\top \Xg^{(i)}} \right\} \,.
\end{align*}
We can apply Assumption \ref{ass:glm} and Lemma \ref{lem:strong-convex} in Appendix \ref{app:lemmas} to bound the second and third term as
\begin{align*}
A\bp{ \bp{\delta + \hbetap}^\top \Xg^{(i)}} - A\bp{ \bp{\betag}^\top \Xg^{(i)}} &\geq A'\bp{\bp{\betag}^\top \Xg^{(i)}} \bp{\delta - \tdelta}^\top \Xg^{(i)} \\
&\hspace{.2in}+ \frac{m}{2} \bn{ \Xg^{(i)} \bp{\delta  - \tdelta } }_2^2 \,.
\end{align*}
Substituting the above yields
\begin{align*}
\cE(\delta) &\geq \frac{1}{\ngold} \sum_{i=1}^{\ngold} \frac{m}{2} \bn{ \Xg^{(i)} \bp{\delta  - \tdelta } }_2^2 \\
&\geq \frac{m}{2 \ngold} \bn{ \Xg \bp{\delta  - \tdelta } }_2^2  \,.
\end{align*}
\Halmos
\endproof

\begin{lemma}\label{lem:w-bound}
The empirical process can be bounded as
\[ \left| w(\hdelta) - w(\tdelta) \right| ~\leq~ \frac{1}{\ngold} \bn{\hdelta - \tdelta}_1 \cdot \bn{\epg^\top \Xg}_{\infty} \,.\]
\end{lemma}

\proof{Proof of Lemma \ref{lem:w-bound}}
First, expanding the loss function and substituting $\tdelta + \hbetap = \betag$ yields
\begin{align*}
\cL(\hdelta) - \cL(\tdelta) = \frac{1}{\ngold} \sum_{i=1}^{\ngold} \bc{-\Yg^{(i)} \bp{\hdelta - \tdelta}^\top \Xg^{(i)} + A\bp{ \bp{\hdelta + \hbetap}^\top \Xg^{(i)}} - A\bp{ \bp{\betag}^\top \Xg^{(i)}}} \,.
\end{align*}
Second, noting that $\E_{\epg} \Yg^{(i)} = A'\bp{ \bp{\betag}^\top \Xg^{(i)}}$, we can write
\begin{align*}
\E_{\epg} \bb{\cL(\hdelta) - \cL(\tdelta)} = \frac{1}{\ngold} \sum_{i=1}^{\ngold}& \left\{-A'\bp{ \bp{\betag}^\top \Xg^{(i)}} \bp{\hdelta - \tdelta}^\top \Xg^{(i)} \right. \\
&\hspace{0.2in} \left. + A\bp{ \bp{\hdelta + \hbetap}^\top \Xg^{(i)}} + A\bp{ \bp{\betag}^\top \Xg^{(i)}} \right\} \,.
\end{align*}
Combining these expressions, we get
\begin{align*}
\left| w(\hdelta) - w(\tdelta) \right| &= \left| \frac{1}{\ngold} \sum_{i=1}^{\ngold} \bc{\bp{-\Yg^{(i)} + A'\bp{ \bp{\betag}^\top \Xg^{(i)}}} \bp{\hdelta - \tdelta}^\top \Xg^{(i)}} \right| \\
&= \left| \bp{\hdelta - \tdelta}^\top \bp{\sum_{i=1}^{\ngold} \Xg^{(i)} \epg^{(i)}} \right|\\
&= \left| \bp{\hdelta - \tdelta}^\top \Xg^\top \epg \right| \\
&\leq \frac{1}{\ngold} \bn{\hdelta - \tdelta}_1 \cdot \bn{\epg^\top \Xg}_{\infty}  \,.
\end{align*}
\Halmos
\endproof

The next lemma is the generalized linear model analog of our earlier Lemma \ref{lem:bound-nu}, and relies on an argument made in Theorem 1 of \cite{chen1999strong}.

\begin{lemma} \label{lem:bound-nu-glm}
On the event $\I$, we have that both
\begin{align*}
\bn{\Xg\nu}_2^2 \le \frac{d\ngold}{m^2\psi^2\nproxy^2}\lambda_1 \,,\quad\quad \text{and} \quad\quad
\bn{\nu}_1 \leq \frac{\sqrt{d\lambda_1}}{m\psi \nproxy} \,.
\end{align*}
\end{lemma}

\proof{Proof of Lemma \ref{lem:bound-nu-glm}}
Recall the generalized linear model's maximum likelihood equation 
\begin{align*}
\hbetap = \arg\min_\beta \sum_{i=1}^{\nproxy} \bc{ -\Yp^{(i)} \beta^\top \Xp^{(i)}  + A\bp{\beta^\top \Xp^{(i)}} - B\bp{\Yp^{(i)} } } \,.
\end{align*}
The first-order optimality condition for $\hbetap$ yields
\begin{align*}
\sum_{i=1}^{\nproxy} \Xp^{(i)} \bp{\Yp^{(i)} - A'\bp{\hbetap^\top \Xp^{(i)}}} &= 0 \,.
\end{align*}
Substituting $\Yp^{(i)} = \mu\bp{{\betap}^\top \Xp^{(i)}} + \epp^{(i)}$ and $A' = \mu$, we get
\begin{align} \label{lem:nu-glm}
\sum_{i=1}^{\nproxy} \Xp^{(i)} \bp{\mu\bp{{\hbetap}^\top \Xp^{(i)}} - \mu\bp{{\betap}^\top \Xp^{(i)}}} &~=~ \sum_{i=1}^{\nproxy} \Xp^{(i)} \epp^{(i)} ~=~ \Xp^\top \epp \,.
\end{align}
Now, note that by applying the mean value theorem, that there exists some $\beta_0$ on the line segment between $\betap$ and $\hbetap$ satisfying
\begin{align*}
\mu\bp{{\hbetap}^\top \Xp^{(i)}} - \mu\bp{{\betap}^\top \Xp^{(i)}} &= \mu'\bp{\beta_0^\top \Xp^{(i)}} \cdot \bp{\hbetap - \betap}^\top \Xp^{(i)} \,.
\end{align*}
Recall that $\nu = \hbetap - \betap$. Substituting the previous expression into Eq. \eqref{lem:nu-glm} and employing a trick from \cite{chen1999strong}, we can write
\begin{align*}
\bn{ \bp{\Xp^\top \Xp}^{-1} \Xp^\top \epp}_2^2 &= \bn{ \bp{\Xp^\top \Xp}^{-1} \sum_{i=1}^{\nproxy} \mu'\bp{\beta_0^\top \Xp^{(i)}} \cdot \Xp^{(i)} {\Xp^{(i)}}^\top \bp{\hbetap - \betap} }_2^2 \\
&=  \nu^\top \bp{\sum_{i=1}^{\nproxy} \mu'\bp{\beta_0^\top \Xp^{(i)}} \cdot \Xp^{(i)} {\Xp^{(i)}}^\top }\bp{\sum_{i=1}^{\nproxy} \Xp^{(i)} {\Xp^{(i)}}^\top}^{-2} \\
&\quad\quad\quad  \times \bp{\sum_{i=1}^{\nproxy} \mu'\bp{\beta_0^\top \Xp^{(i)}} \cdot \Xp^{(i)} {\Xp^{(i)}}^\top } \nu \\
& \geq \nu^\top \bp{\sum_{i=1}^{\nproxy} \mu'\bp{\beta_0^\top \Xp^{(i)}} \cdot \Xp^{(i)} {\Xp^{(i)}}^\top }\bp{\sum_{i=1}^{\nproxy} \frac{\mu'\bp{\beta_0^\top \Xp^{(i)}}}{m} \Xp^{(i)} {\Xp^{(i)}}^\top}^{-2} \\
&\quad\quad\quad \times \bp{\sum_{i=1}^{\nproxy} \mu'\bp{\beta_0^\top \Xp^{(i)}} \cdot \Xp^{(i)} {\Xp^{(i)}}^\top } \nu \\
&= m^2 \bn{ \nu}^2_2 \,,
\end{align*}
where we have used the fact that $\mu'\bp{z} \geq m > 0$ for all $z$ in the domain since $A$ is strongly convex (see Lemma \ref{lem:strong-convex}). Thus, we have
\[ \bn{\nu}_2 \leq \frac{1}{m} \bn{ \bp{\Xp^\top \Xp}^{-1} \Xp^\top \epp}_2 \,.\]
We now proceed exactly as we did in Lemma \ref{lem:bound-nu} in the linear case; we omit several algebraic details to avoid repetition.
For the first inequality, on event $\I$, we can write
\begin{align*}
\bn{\Xg\nu}_2^2 &\leq \frac{1}{m^2} \bn{\Xg(\Xp^\top\Xp)^{-1}\Xp^\top\epp}_2^2 \\
&\leq \frac{1}{m^2} \bn{\Xg}_2^2\cdot\bn{(\Xp^\top\Xp)^{-1}\Xp^\top\epp}_2^2 \\
&\le \frac{d\ngold}{m^2\psi^2\nproxy^2}\lambda_1 \,.
\end{align*}
For the second inequality, we observe that on event $\I$,
\begin{align*}
\bn{\nu}_1 &\leq \sqrt{d} \bn{\nu}_2 \\
&\leq \frac{\sqrt{d }}{m}\bn{(\Xp^\top\Xp)^{-1} }_2 \cdot \bn{ \Xp^\top\epp}_2 \\
&\leq \frac{\sqrt{d\lambda_1}}{m\psi \nproxy} \,.
\end{align*}
\Halmos
\endproof

\subsection{Proof of Lemma \ref{lem:basic-ineq-glm}}

\begin{lemma}\label{lem:basic-ineq-glm}
On the event $\J$, taking $\lambda \geq 5 \lambda_0/2$, the solution $\hdelta$ to the optimization problem \eqref{eq:obj-glm2} satisfies
\begin{align*}
\lambda \bn{ \tdelta - \hdelta }_1 ~&\leq~ \frac{5m}{8\ngold}\bn{ \Xg \nu }_2^2 + \frac{5\lambda^2 s}{2m \phi^2} + 5\lambda \bn{ \nu }_1 \,.
\end{align*}
\end{lemma}

\proof{Proof of Lemma \ref{lem:basic-ineq-glm}}
Since the optimization problem \eqref{eq:obj-glm2} is convex, it recovers the in-sample global minimum. Thus, we must have that
\[\cL(\hdelta;~\Xg, \Yg) + \lambda \bn{\hdelta}_1 ~\leq~ \cL(\tdelta;~\Xg, \Yg) + \lambda \bn{\tdelta}_1 \,.\]
We can re-write this as
\begin{align}
\cE(\hdelta) + \lambda\bn{\hdelta}_1 &\leq -\bb{w(\hdelta) - w(\tdelta)} + \lambda\bn{\tdelta}_1 \nonumber \\
&\leq \frac{1}{\ngold} \bn{\hdelta - \tdelta}_1 \cdot \bn{\epg^\top \Xg}_{\infty} + \lambda\bn{\tdelta}_1 \nonumber \\
&\leq \frac{\lambda}{5} \bn{\hdelta - \tdelta}_1 + \lambda\bn{\tdelta}_1 \nonumber \\
&= \frac{\lambda}{5} \bn{\hdelta - \delta^* + \nu}_1 + \lambda\bn{\delta^* -\nu}_1 \,, \label{eq:basic-glm-1}
\end{align}
where we have used Lemma \ref{lem:w-bound} in the second inequality, the fact that $\J$ holds and that $\lambda \geq 5\lambda_0/2$ in the third inequality, and the fact that $\nu = \delta^*-\tdelta$ in the last equality. Now, similar to the linear case (see Lemma \ref{lem:basic-ineq}), we use the triangle inequality to express $\hdelta$ in terms of its components on the index set $S$. Collecting Eqs. \eqref{eq:triangle-1}--\eqref{eq:triangle-2} and substituting into Eq. \eqref{eq:basic-glm-1}, we obtain
\begin{align*}
5\cE(\hdelta) + 5\lambda\bp{\bn{\delta_S^* }_1 - \bn{ \hdelta_S - \delta_S^* }_1 +  \bn{\hdelta_{S^c} }_1} &~\leq~ \lambda \bp{\bn{ \hdelta_S - \delta_S^* }_1 + \bn{\hdelta_{S^c} }_1 + \bn{ \nu }_1 } +5 \lambda\bp{\bn{ \delta^* }_1 + \bn{ \nu }_1} \,.
\end{align*}
Cancelling terms on both sides yields
\begin{align}
5\cE(\hdelta) + 4\lambda \bn{\hdelta_{S^c} }_1 &~\leq~ 6\lambda \bp{\bn{ \hdelta_S - \delta_S^* }_1 + \bn{ \nu }_1 }\,. \label{eq:basic-glm-2}
\end{align}
As in the linear case, we have two possible cases: either (i) $\bn{ \nu }_1 \leq \bn{\hdelta_S - \delta_S^* }_1$, or (ii) $\bn{\hdelta_S - \delta_S^* }_1 < \bn{ \nu }_1 $. In Case (i), we will invoke the compatibility condition to prove our finite-sample guarantee for the joint estimator, and in Case (ii), we will find that we already have good control over the error of the estimator.

\textbf{Case (i):} We are in the case that $\bn{ \nu }_1 \leq \bn{\hdelta_S - \delta_S^* }_1$, so from Eq. (\ref{eq:basic-glm-2}), we can write on $\J$,
\[ 5\cE(\hdelta) + 4\lambda \bn{\hdelta_{S^c} }_1 ~\leq~ 12 \lambda \bn{\hdelta_S - \delta_S^* }_1 \,. \]
Dropping the first (non-negative) term on the left hand side, we immediately observe that on $\J$,
\[ \bn{\hdelta_{S^c} }_1 = \bn{\hdelta_{S^c}  - \delta^*_{S^c}}_1 ~\leq~ 3 \bn{\hdelta_S - \delta_S^* }_1 \,,\]
so we can apply the compatibility condition (Definition \ref{def:cc}) to $u = \hdelta - \delta^*$ and take the square root. This yields
\[ \bn{\hdelta_S  - \delta^*_S}_1 ~\leq~ \frac{\sqrt{s}}{\phi \sqrt{\ngold}} \bn{ \Xg \bp{\hdelta  - \delta^* } }_2 \,. \]
Separately, when Case (i) and $\J$ hold, we can further simplify
\begin{align} \label{eq:basic-glm-3}
5\cE(\hdelta) + 4 \lambda \bn{ \tdelta - \hdelta }_1 ~&=~ 5\cE(\hdelta) + 4\lambda \bn{\hdelta - \delta^* + \nu }_1 \nonumber \\
&\leq~ 5\cE(\hdelta) + 4\lambda \bn{\hdelta_S - \delta^*_S}_1 + 4\lambda \bn{ \hdelta_{S^c} }_1 + 4 \lambda \bn{\nu }_1 \nonumber \\
&\leq~ 10 \lambda \bn{ \hdelta_S - \delta^* }_1 + 10 \lambda \bn{\nu }_1 \nonumber \\
&\leq \frac{10 \lambda \sqrt{s}}{\phi \sqrt{\ngold}} \bn{ \Xg \bp{\hdelta  - \delta^* } }_2 + 10 \lambda \bn{ \nu }_1 \,.
\end{align}
where we used Eq. (\ref{eq:basic-glm-2}) in the second inequality, and the compatibility condition to bound $\bn{ \hdelta_S - \delta^* }_1$ in the third inequality.

We now need to relate the expected error relative to the true minimizer $\cE(\hdelta)$ to $\bn{ \Xg \bp{\hdelta  - \delta^* } }_2$ in order to (partially) cancel the term $\frac{10 \lambda \sqrt{s}}{\phi \sqrt{\ngold}} \bn{ \Xg \bp{\hdelta  - \delta^* } }_2$ on the right hand side. In the linear case, these quantities are trivially related since $\cE(\delta) = \frac{1}{\ngold}\bn{ \Xg \bp{\hdelta  - \delta^* } }_2^2$. When considering more general nonlinear loss functions, one needs to additionally impose a margin condition and a suitable alternative compatibility condition in order to establish a relationship between these two terms \citep[see, e.g.,][]{negahban, buhlmann}. However, this additional infrastructure is not necessary for generalized linear models due to their close connection to linear models. Lemma \ref{lem:E-bound} shows that it is sufficient to use the functional form of the GLM log likelihood and the strong convexity of $A(\cdot)$ to establish a relationship.

Applying Lemma \ref{lem:E-bound}, we can write
\begin{align*}
\frac{5m}{2\ngold} \bn{ \Xg \bp{\hdelta  - \tdelta } }_2^2 + 4 \lambda \bn{ \tdelta - \hdelta }_1 ~&\leq~ 5\cE(\hdelta) + 4 \lambda \bn{ \tdelta - \hdelta }_1 \\
&\leq \frac{10 \lambda \sqrt{s}}{\phi \sqrt{\ngold}} \bn{ \Xg \bp{\hdelta  - \delta^* } }_2 + 10 \lambda \bn{ \nu }_1 \\
&\leq  \frac{5m}{4\ngold} \bn{ \Xg \bp{\hdelta  - \delta^* } }_2^2 + \frac{20\lambda^2 s}{m \phi^2} + 10 \lambda \bn{ \nu }_1 \\
&\leq \frac{5m}{2\ngold} \bn{ \Xg \bp{\hdelta  - \tdelta } }_2^2 + \frac{5m}{2\ngold} \bn{ \Xg \nu }_2^2 + \frac{20\lambda^2 s}{m \phi^2} + 10 \lambda \bn{ \nu }_1\,,
\end{align*}
where we have used Eq. \eqref{eq:basic-glm-3} in the second inequality, the identity that $2ab \leq a^2 + b^2$ for  
\[ a = \sqrt{\frac{5m}{4\ngold}}\bn{ \Xg \bp{\hdelta  - \delta^* } }_2 \quad\quad \text{and} \quad\quad b = \frac{\lambda}{\phi}\sqrt{\frac{20s}{m}} \,, \]
in the third inequality, and the identity that $(a+b)^2 \leq 2a^2+2b^2$ in the last inequality. Cancelling terms on both sides, we find that when $\J$ and Case (i) hold, we have that
\begin{align} \label{eq:case1-glm}
\lambda \bn{ \tdelta - \hdelta }_1 ~&\leq~ \frac{5m}{8\ngold}\bn{ \Xg \nu }_2^2 + \frac{5\lambda^2 s}{m \phi^2} + \frac{5\lambda \bn{ \nu }_1 }{2} \,.
\end{align}

\textbf{Case (ii):} We are in the case that $\bn{\hdelta_S - \delta_S^* }_1 \leq \bn{ \nu }_1$, so Eq. (\ref{eq:basic-glm-2}) implies on $\J$,
\begin{align*}
5 \cE(\hdelta) + 4 \lambda \bn{\hdelta_{S^c} }_1 ~\leq~ 12 \lambda \bn{\nu }_1 \,.
\end{align*}
In this case, we do not actually need to invoke the compatibility condition. When $\J$ and Case (ii) hold, we can directly bound
\begin{align*} 
5 \cE(\hdelta) + 4 \lambda \bn{ \tdelta - \hdelta }_1 ~&\leq~ 5 \cE(\hdelta) + 4\lambda \bn{\hdelta_S - \delta^*_S}_1 + 4\lambda \bn{ \hdelta_{S^c} }_1 + 4 \lambda \bn{\nu }_1 \\
&\leq~ 20 \lambda \bn{ \nu }_1 \,,
\end{align*}
where we used the triangle inequality (see Eq. \eqref{eq:triangle-2} from the proof in the linear case) in the first inequality and Eq. \eqref{eq:case2-int} as well as the fact that $\bn{\hdelta_S - \delta_S^* }_1 \leq \bn{ \nu }_1$ in the second inequality. Dropping the first (non-negative) term on the left hand side yields
\begin{align} \label{eq:case2-glm}
\lambda \bn{ \tdelta - \hdelta }_1 ~&\leq~ 5 \lambda \bn{ \nu }_1 \,.
\end{align}
Combining the inequalities from Eq. (\ref{eq:case1-glm}) and (\ref{eq:case2-glm}), the following holds in both cases on $\J$,
\begin{align} \label{eq:combine-glm}
\lambda \bn{ \tdelta - \hdelta }_1 ~\leq~ \frac{5m}{8\ngold}\bn{ \Xg \nu }_2^2 + \frac{5\lambda^2 s}{m \phi^2} + 5\lambda \bn{ \nu }_1  \,.
\end{align}
\Halmos
\endproof

\subsection{Proof of Theorem \ref{thm:glm} and Corollary \ref{cor:glm}} \label{app:glm-thm}

Having established Lemma \ref{lem:basic-ineq-glm}, the proof of this theorem follows the proof in the linear case closely. Note that we have defined the same high probability events $\J$ and $\I$ for the generalized linear model setting, and thus these events satisfy the same tail inequalities, allowing us to re-use Lemmas \ref{lem:J} and \ref{lem:I}.

\proof{Proof of Theorem \ref{thm:glm}} 
By Lemma \ref{lem:basic-ineq-glm}, we can bound $\bn{ \tdelta - \hdelta }_1$ with high probability on $\J$ when $\nu$ is small. We can re-use Lemma \ref{lem:bound-nu} from the linear case to bound the terms that depend on $\nu$ as a function of $\nproxy$ on the event $\I$. Applying Lemma \ref{lem:bound-nu-glm} to Lemma \ref{lem:basic-ineq-glm} yields
\begin{align*}
 \bn{ \tdelta - \hdelta }_1 ~\leq~ \frac{5d \lambda_1}{8m\psi^2\nproxy^2 \lambda} + \frac{5\lambda s}{m \phi^2} + \frac{5\sqrt{d \lambda_1}}{m\psi \nproxy}  \,.
\end{align*}
The above expression holds with probability $1$ when the events $\J$ and $\I$ hold, and $\lambda \geq 5\lambda_0/2$. Recall that $\lambda_0, \lambda_1$ are theoretical quantities that we can choose freely to optimize our bound. In contrast, $\lambda$ is a fixed regularization parameter chosen by the decision-maker when training the estimator. Then, setting $\lambda_0 = 2\lambda/5$, we can write
\begin{align*}
\Pr\bb{\bn{ \tdelta - \hdelta }_1 ~\geq~ \frac{5d \lambda_1}{8m\psi^2\nproxy^2 \lambda} + \frac{5\lambda s}{m \phi^2} + \frac{5\sqrt{d \lambda_1}}{m\psi \nproxy}} &~\leq~ 1 - \Pr[\J \cap \I] \\
&~\leq~ \Pr[\J^C] + \Pr[\I^C] \\
&~\leq~ 2d \exp\bp{-\frac{\lambda^2 \ngold}{50 \sigmag^2}} + 2d \exp\bp{-\frac{\lambda_1}{2 d \sigmap^2 \nproxy}} \,.
\end{align*}
The second inequality follows from a union bound, and the third follows from Lemma \ref{lem:J} (setting $\lambda_0 = 2\lambda/5$) and Lemma \ref{lem:I}.
By inspection, we choose
\[ \lambda_1 = \frac{\nproxy^2 \lambda^2}{d} \,,\]
yielding
\begin{align*}
\Pr\bb{\bn{ \tdelta - \hdelta }_1 ~\geq~ \frac{5\lambda}{m} \bp{\frac{1}{8\psi^2} + \frac{1}{\psi} + \frac{s}{\phi^2} }} &~\leq~ 2d \exp\bp{-\frac{\lambda^2 \ngold}{50 \sigmag^2}} + 2d \exp\bp{-\frac{\lambda^2 \nproxy}{2 d^2 \sigmap^2}} \,.
\end{align*}
Finally, we reverse our variable transformation by substituting $\hbeta_{joint} = \hdelta + \hbetap$ and $\betag = \tdelta + \hbetap$, which gives us the result.
\Halmos
\endproof

The following corollary uses the tail inequality in Theorem \ref{thm:glm} to obtain an upper bound on the expected parameter estimation error of the truncated joint estimator.
\begin{corollary}[Joint Estimator] \label{cor:glm} The parameter estimation error of the truncated nonlinear joint estimator for a generalized linear model is bounded above as follows:
\begin{align*}
R \bp{ \hbeta_{joint}^{tr}(\lambda), \betag } ~&\leq~ \frac{5\lambda}{m} \bp{\frac{1}{8\psi^2} + \frac{1}{\psi} + \frac{s}{\phi^2}} + 6bd \bp{\exp\bp{-\frac{\lambda^2 \ngold}{50 \sigmag^2}} + \exp\bp{-\frac{\lambda^2 \nproxy}{2 d^2 \sigmap^2}}}\,.
\end{align*}
Let $C >0$ be any tuning constant. Taking the regularization parameter to be
\begin{align*}
\blambda ~&=~ \max\bc{\sqrt{\frac{50 \sigmag^2 \log\bp{6bd\ngold}}{\ngold}},~ \sqrt{\frac{2d^2 \sigmap^2 \log\bp{6bd\nproxy}}{\nproxy}}} =~ \tO\bp{\frac{\sigmag }{\sqrt{\ngold}} + \frac{d \sigmap }{ \sqrt{\nproxy}}} \,,
\end{align*}
yields a parameter estimation error of order
\begin{align*}
R \bp{ \hbeta_{joint}^{tr}(\blambda), \betag } ~&=~ \tO\bp{\frac{s \sigmag }{\sqrt{\ngold}} + \frac{sd \sigmap }{\sqrt{\nproxy}}} \,.
\end{align*}
\end{corollary}

\proof{Proof of Corollary \ref{cor:glm}} For ease of notation, let 
\[ w = \frac{5}{m} \bp{\frac{1}{8\psi^2} + \frac{1}{\psi} + \frac{s}{\phi^2}}   \,.\]
Recall from the proof of Theorem \ref{thm:glm} that
\[ \bn{ \hbeta_{joint}(\lambda) - \betag  }_1 ~\leq~  \lambda w\,,\]
with probability $1$ when the events $\J$ and $\I$ hold, and we take $\lambda_0 = 2\lambda/5$ and $\lambda_1 = \nproxy^2 \lambda^2/d$.

From here, we can use the proof of Corollary \ref{cor:joint} exactly, and thus avoid repeating the steps.
Collecting Eqs. \eqref{eq:cor-first} and \eqref{eq:cor-second} and substituting into Eq. \eqref{eq:cor-exp} from the proof in the linear case, we have
\begin{align*}
\E\bb{\bn{\hbeta_{joint}^{tr}(\lambda) - \betag}_1} ~&\leq~ \lambda w \cdot \Pr[\J \cap \I] + 3b \cdot \Pr[\J^C \cup \I^C] \\
&\leq \lambda w + 3b \cdot \bp{\Pr[J^C] + \Pr[\I^C] } \\
&\leq \lambda w + 6bd \cdot \bp{\exp\bp{-\frac{\lambda^2 \ngold}{50 \sigmag^2}} + \exp\bp{-\frac{\lambda^2 \nproxy}{2 d^2 \sigmap^2}}}\,,
\end{align*}
using a union bound, and applying Lemmas \ref{lem:J} and \ref{lem:I} with the chosen values $\lambda_0 = 2\lambda/5$ and $\lambda_1 = \nproxy^2 \lambda^2/d$.
\Halmos
\endproof

\section{$\ell_2$ error of Joint Estimator} \label{app:l2}

We now extend Theorem \ref{thm:main} to prove parameter estimation bounds in the $\ell_2$ norm. We follow the general approach outlined in prior work \citep{candes, bickel, buhlmann}, while additionally accounting for our problem's approximate sparsity. As noted in prior work, for this setting, we need to replace the compatibility condition (Assumption \ref{ass:cc}) with a stronger assumption known as the restricted eigenvalue condition (Assumption \ref{ass:cc2}) defined below:

\begin{definition}[Restricted Eigenvalue Condition] \label{def:cc2} For any index set $\cN \supset S$ with $|\cN| = N$, let
\[ \mathfrak{R}(S,\cN)= \bc{u \in \R^d:~ \| u_{S^c} \|_1 \leq 3 \|u_S\|_1,~ \|u_{\cN^c} \|_{\infty} \leq \min_{j \in \cN \setminus S} |u_j|} \,.\]
The $(S,N)$-restricted eigenvalue condition is met for the matrix $\Sigma \in \R^{d \times d}$ with constant $\phi_R > 0$, if for every $\cN \supset S$ with $|\cN| = N$, and for all $u \in \mathfrak{R}(S,\cN)$, it holds that
\[ \|u_{\cN}\|_2^2 \leq \bp{u^T \Sigma u}/\phi_R^2 \,. \]
\end{definition}

\begin{assumption}[Restricted Eigenvalue Condition] \label{ass:cc2} The restricted eigenvalue condition (Definition \ref{def:cc2}) is met for $S = supp(\delta^*)$, $N = 2s$, and the gold sample covariance matrix $\Sigmag$ with constant $\phi_R > 0$.
\end{assumption}

Then, under Assumptions \ref{ass:bounded}, \ref{ass:proxy-posdef} and \ref{ass:cc2}, we have the following tail inequality that upper bounds the $\ell_2$ parameter estimation error of the two-step joint estimator with high probability.
\begin{theorem}[Joint Estimator] \label{thm:rev} The joint estimator satisfies the following tail inequality for any chosen value of the regularization parameter $\lambda > 0$:
\[ \Pr\bb{\bn{ \hbeta_{joint}(\lambda) - \betag  }_2 ~\geq~ \lambda \bp{\frac{25}{16 \sqrt{s} \psi^2 } + \frac{5}{\psi } + \frac{21 \sqrt{s}}{\phi_R^2}}} ~\leq~ 2d \exp\bp{-\frac{\lambda^2 \ngold}{200 \sigmag^2}} + 2d \exp\bp{-\frac{\lambda^2 \nproxy}{2 d^2 \sigmap^2}} \,. \]
\end{theorem}
This result is identical to the $\ell_1$ bound provided in Theorem \ref{thm:main} (with modified constants), with the exception that the $\ell_2$ bound scales as $\sqrt{s}$ rather than $s$. Specifically, taking the regularization parameter to be
\begin{align*}
\blambda ~&=~ C\max\bc{\sqrt{\frac{200 \sigmag^2 \log\bp{6bd\ngold}}{\ngold}},~ \sqrt{\frac{2d^2 \sigmap^2 \log\bp{6bd\nproxy}}{\nproxy}}} =~ \tO\bp{\frac{\sigmag }{\sqrt{\ngold}} + \frac{d \sigmap }{ \sqrt{\nproxy}}} \,,
\end{align*}
one can easily verify that the worst-case $\ell_2$ parameter estimation error is of order
\begin{align*}
\sup_{\mathcal{S}} ~\E\bb{\bn{\hbeta_{joint} - \betag}_2} ~&=~ \cO\bp{\max\bc{ \frac{\sqrt{s} \sigmag }{\sqrt{\ngold}} \log\bp{d\ngold} ,~ \frac{\sqrt{s}d \sigmap }{\sqrt{\nproxy}}\log\bp{d\nproxy}}} \,.
\end{align*}
This scaling is consistent with the LASSO literature \citep[see, e.g.,][]{bickel}. 

\subsection{Proof of Lemma \ref{lem:rev-1}}

We first prove an analog of Lemma \ref{lem:basic-ineq}, omitting repeated steps.

\begin{lemma} \label{lem:rev-1}
On the event $\J$, taking $\lambda \geq 5 \lambda_0$, the solution $\hdelta$ to the optimization problem (\ref{eq:obj2}) satisfies
\[ \bn{ \tdelta - \hdelta }_2 ~\leq~ \frac{21 \lambda \sqrt{s}}{\phi_R^2} + 5 \bn{ \nu }_1 + \bp{\frac{3}{\phi_R\sqrt{\ngold}} + \frac{5}{4\lambda s \ngold}} \bn{ \Xg \nu}_2 \,. \]
\end{lemma}

\proof{Proof of Lemma \ref{lem:rev-1}}
Following the same steps as Lemma \ref{lem:basic-ineq}, we obtain Eq. \eqref{eq:basic-on-J} on $\J$:
\begin{align*}
\frac{5}{\ngold} \bn{\Xg\bp{\tdelta - \hdelta} }^2_2 + 5 \lambda \bn{ \hdelta }_1 ~&\leq~ \lambda \bn{\hdelta - \delta^* + \nu }_1 + 5\lambda \bn{ \delta^* - \nu }_1 \,.
\end{align*}
Next, we express $\hdelta$ in terms of its components on a new index set $\cN$ rather than $S$. Define the indices $\{j_1, \cdots, j_{d-s}\} \subset [d]\setminus S$ such that $|\hdelta_{j_1}| > |\hdelta_{j_2}| > \cdots > |\hdelta_{j_{d-s}}|$. Then, let the index set
\[ \cN = S \cup \{j_1, \cdots j_s \} \,. \]
In words, $\cN$ contains $S$ and the indices corresponding to the $s$ largest (in absolute value) components of $\hdelta_{S^c}$. 

Once again, by the triangle inequality, we have
\begin{align}
\bn{ \hdelta }_1 ~&=~ \bn{\hdelta_{\cN} }_1 + \bn{ \hdelta_{\cN^c} }_1 ~\geq~ \bn{\delta_\cN^* }_1 - \bn{ \hdelta_\cN - \delta_\cN^* }_1 +  \bn{\hdelta_{\cN^c} }_1 \,. \label{eq:triangle-1-rev}
\end{align}
Similarly, noting that $\delta_{\cN^c}^* = 0$ since $\cN \supset S$, we have
\begin{align}
\bn{\hdelta - \delta^* + \nu }_1 ~&\leq~ \bn{ \hdelta_{\cN} - \delta_{\cN}^* }_1 + \bn{\hdelta_{\cN^c} }_1 + \bn{ \nu }_1 \,. \label{eq:triangle-2-rev}
\end{align}
Substituting Eq. \eqref{eq:triangle-1-rev}--\eqref{eq:triangle-2-rev} into Eq. (\ref{eq:basic-on-J}), we have that when $\J$ holds,
\begin{equation} \label{eq:basic-on-J-2-rev}
\frac{5}{\ngold} \bn{\Xg \bp{\tdelta - \hdelta} }^2_2 + 4 \lambda \bn{\hdelta_{\cN^c} }_1 ~\leq~ 6 \lambda \bn{\hdelta_{\cN} - \delta_{\cN^*} }_1 + 6\lambda \bn{ \nu }_1 \,.
\end{equation}
Again, we proceed by considering two cases: either (i) $\bn{ \nu }_1 \leq \bn{\hdelta_{\cN} - \delta_{\cN^*} }_1$, or (ii) $\bn{\hdelta_{\cN} - \delta_{\cN^*} }_1 < \bn{ \nu }_1 $.

\textbf{Case (i):} We are in the case that $\bn{ \nu }_1 \leq \bn{\hdelta_{\cN} - \delta_{\cN^*} }_1$. Applying this inequality and dropping the first (non-negative) term on the left hand side of Eq. (\ref{eq:basic-on-J-2-rev}), we observe that
\[ \bn{\hdelta_{\cN^c} }_1 = \bn{\hdelta_{\cN^c}  - \delta^*_{\cN^c}}_1 ~\leq~ 3 \bn{\hdelta_{\cN} - \delta_{\cN^*} }_1 \,.\]
Furthermore, by the construction of $\cN$, $\max_{j \in \cN^c}|\hdelta_j| \leq \min_{j \in \cN \setminus S} |\hdelta_j|$, which implies that $\|\hdelta_{\cN^c}  - \delta^*_{\cN^c} \|_{\infty} \leq \min_{j \in \cN \setminus S} |\hdelta_j - \delta^*_j|$. Thus, we can apply the restricted eigenvalue condition to $u = \hdelta - \delta^*$, yielding
\begin{equation} \label{eq:case1-int-rev}
\bn{\hdelta_{\cN}  - \delta^*_{\cN}}^2_2 ~\leq~ \frac{1}{\phi_R^2 \ngold} \bn{ \Xg \bp{\hdelta  - \delta^* } }_2^2 \,.
\end{equation}
Furthermore, since $\cN$ only contains $2s$ indices, note that $\bn{\hdelta_{\cN}  - \delta^*_{\cN}}_1 \leq \sqrt{2s} \bn{\hdelta_{\cN}  - \delta^*_{\cN}}_2$. Then, when Case (i) and $\J$ hold, we can apply Eq. \eqref{eq:triangle-2-rev}, \eqref{eq:basic-on-J-2-rev} and \eqref{eq:case1-int-rev} to simplify
\begin{align*}
\frac{5}{\ngold} \bn{ \Xg \bp{\hdelta - \delta^*} }_2^2 + 4 \lambda \bn{ \tdelta - \hdelta }_1 ~&\leq~ \frac{10}{\ngold} \bn{ \Xg \bp{\tdelta - \hdelta} }_2^2 +  \frac{10}{\ngold} \bn{ \Xg \nu}_2^2 + 4 \lambda \bn{ \hdelta - \delta^* + \nu }_1 \\
&\leq~ \frac{10}{\ngold} \bn{ \Xg \bp{\tdelta - \hdelta} }_2^2 +  \frac{10}{\ngold} \bn{ \Xg \nu}_2^2  + 4\lambda \bn{\hdelta_{\cN} - \delta^*_{\cN}}_1 + 4\lambda \bn{ \hdelta_{\cN^c} }_1 \\
&\hspace{0.2in}+ 4 \lambda \bn{\nu }_1 \\
&\leq~ 16 \lambda \bn{ \hdelta_{\cN} - \delta_{\cN}^* }_1 + \frac{10}{\ngold} \bn{ \Xg \nu}_2^2 + 16 \lambda \bn{\nu }_1 \\
&\leq~ 32 \lambda \bn{ \hdelta_{\cN} - \delta_{\cN}^* }_1 + \frac{10}{\ngold} \bn{ \Xg \nu}_2^2 \\
&\leq~ \frac{32 \lambda \sqrt{2s}}{\phi_R \sqrt{\ngold}} \bn{ \Xg \bp{\hdelta  - \delta^* } }_2 + \frac{10}{\ngold} \bn{ \Xg \nu}_2^2 \\
&\leq~ \frac{3}{\ngold} \bn{ \Xg \bp{\hdelta - \delta^*} }_2^2 + \frac{512 \lambda^2 s}{\phi_R^2} + \frac{10}{\ngold} \bn{ \Xg \nu}_2^2 \,,
\end{align*}
where the first and sixth inequalities follow from the facts that $(a+b)^2 \leq 2a^2 + 2b^2$ and $32\sqrt{2}ab \leq 3a^2 + 512b^2/3$ for any $a,b \in \R$, the third inequality follows from Eq. \eqref{eq:basic-on-J-2-rev}, the fourth inequality follows because we are in Case (i), and the fifth inequality follows from the restricted eigenvalue condition in Eq. \eqref{eq:case1-int-rev}. Dropping the second (non-negative) term on the left hand side and simplifying, we get
\begin{align*}
\frac{1}{\ngold} \bn{ \Xg \bp{\hdelta - \delta^*} }_2^2 ~\leq~ \frac{256 \lambda^2 s}{\phi_R^2} + \frac{5}{\ngold} \bn{ \Xg \nu}_2^2 \,.
\end{align*}
Applying the restricted eigenvalue condition in Eq. \eqref{eq:case1-int-rev} again, we can write
\begin{align*}
\bn{ \hdelta_{\cN} - \delta^*_{\cN} }_2^2 ~&\leq~ \frac{1}{\phi_R^2 \ngold} \bn{ \Xg \bp{\hdelta - \delta^*} }_2^2 \\
&\leq~ \frac{256 \lambda^2 s}{\phi_R^4} + \frac{5}{\phi_R^2\ngold} \bn{ \Xg \nu}_2^2 \,.
\end{align*}
Using the fact that $\sqrt{a+b} \leq \sqrt{a} + \sqrt{b}$, we can bound
\begin{align} \label{eq:rev-n-bound}
\lambda \bn{ \hdelta_{\cN} - \delta^*_{\cN} }_2 ~&\leq~ \frac{16 \lambda^2 \sqrt{s}}{\phi_R^2} + \frac{\sqrt{5} \lambda}{\phi_R\sqrt{\ngold}} \bn{ \Xg \nu}_2 \,.
\end{align}
Now, it remains to bound $\|\hdelta_{\cN^c} \|_2$, for which we need to bound $\| \hdelta_{S^c} \|_1$. Employing a similar strategy but different constants, we can write
\begin{align*}
\frac{5}{\ngold} \bn{ \Xg \bp{\tdelta - \hdelta} }_2^2 + 4 \lambda \bn{ \tdelta - \hdelta }_1 ~&\leq~ \frac{5}{\ngold} \bn{ \Xg \bp{\tdelta - \hdelta} }_2^2 + 4\lambda \bn{\hdelta_{\cN} - \delta^*_{\cN}}_1 + 4\lambda \bn{ \hdelta_{\cN^c} }_1 + 4 \lambda \bn{\nu }_1 \\
&\leq~ 10 \lambda \bn{ \hdelta_{\cN} - \delta_{\cN}^* }_1 + 10 \lambda \bn{\nu }_1 \\
&\leq~ \frac{10 \lambda \sqrt{2s}}{\phi_R \sqrt{\ngold}} \bn{ \Xg \bp{\hdelta  - \delta^* } }_2 + 10 \lambda \bn{ \nu }_1 \\
&\leq~ \frac{5}{2\ngold} \bn{ \Xg \bp{\hdelta - \delta^*} }_2^2 + 10 \lambda \bn{ \nu }_1 + \frac{20 \lambda^2 s}{ \phi_R^2} \\
&\leq~ \frac{5}{\ngold} \bn{ \Xg \bp{\tdelta - \hdelta} }_2^2 + \frac{5}{\ngold} \bn{ \Xg \nu }_2^2 + 10 \lambda \bn{ \nu }_1 + \frac{20 \lambda^2 s}{\phi_R^2} \,,
\end{align*}
where the fourth and fifth inequalities follows from the facts that $10\sqrt{2}ab \leq 5a^2/2 + 20b^2$ and $(a+b)^2 \leq 2a^2 + 2b^2$ for any $a,b \in \R$. 
As a result, we can bound
\begin{align*}
\lambda \bn{ \hdelta_{S^c} }_1 ~\leq~ \lambda \bn{ \tdelta - \hdelta }_1 ~\leq~ \frac{5}{4 \ngold} \bn{ \Xg \nu }_2^2 + \frac{5\lambda}{2} \bn{ \nu }_1 + \frac{5 \lambda^2 s}{\phi_R^2} \,.
\end{align*}
Invoking Lemma \ref{lem:ell-q} from Appendix \ref{app:lemmas}, by the construction of $\cN$, we have
\begin{align} \label{eq:rev-nc-bound}
\lambda \bn{ \hdelta_{\cN^c} }_2 ~\leq~ s^{-1/2} \bn{ \hdelta_{S^c} }_1 ~\leq~ \frac{5}{4 \sqrt{s} \ngold} \bn{ \Xg \nu }_2^2 + \frac{5\lambda}{2\sqrt{s}} \bn{ \nu }_1 + \frac{5 \lambda^2 \sqrt{s}}{\phi_R^2} \,.
\end{align}
Then, combining Eq. \eqref{eq:rev-n-bound} and \eqref{eq:rev-nc-bound}, we have that when $\J$ and Case (i) hold,
\begin{align} \label{eq:case1-rev}
\lambda \bn{ \tdelta - \hdelta }_2 ~&\leq~ \lambda \bn{ \hdelta_{\cN} - \delta^*_{\cN} }_2 + \lambda\bn{ \hdelta_{\cN^c} }_2 \nonumber \\
&\leq~ \frac{5}{4 \sqrt{s} \ngold} \bn{ \Xg \nu }_2^2 + \frac{5\lambda}{2\sqrt{s}} \bn{ \nu }_1 + \frac{\sqrt{5} \lambda}{\phi_R\sqrt{\ngold}} \bn{ \Xg \nu}_2 + \frac{21 \lambda^2 \sqrt{s}}{\phi_R^2} \nonumber \\
&\leq~ \frac{25}{16 \sqrt{s} \ngold} \bn{ \Xg \nu }_2^2 + \frac{5\lambda}{2\sqrt{s}} \bn{ \nu }_1 + \frac{21 \lambda^2 \sqrt{s}}{\phi_R^2} \,, 
\end{align}
where the last inequality follows from the fact that $\sqrt{5}ab \leq 5a^2/4 + b^2$ for any $a,b \in \R$.

\textbf{Case (ii):} We are in the case that $\bn{\hdelta_{\cN} - \delta_{\cN}^* }_1 \leq \bn{ \nu }_1$, so Eq. (\ref{eq:basic-on-J-2-rev}) implies on $\J$,
\begin{align*}
\frac{5}{\ngold} \bn{\Xg \bp{\tdelta - \hdelta} }^2_2 + 4 \lambda \bn{\hdelta_{\cN^c} }_1 ~\leq~ 12 \lambda \bn{\nu }_1 \,.
\end{align*}
In this case, we do not actually need to invoke the restricted eigenvalue condition, so we can proceed exactly as we did in Lemma \ref{lem:basic-ineq}. When $\J$ and Case (ii) hold, we can directly bound
\begin{align*} 
\frac{5}{\ngold} \bn{ \Xg \bp{\tdelta - \hdelta} }_2^2 + 4 \lambda \bn{ \tdelta - \hdelta }_1 ~&\leq~ \frac{5}{\ngold} \bn{ \Xg \bp{\tdelta - \hdelta} }_2^2 + 4\lambda \bn{\hdelta_{\cN} - \delta^*_{\cN}}_1 + 4\lambda \bn{ \hdelta_{\cN^c} }_1 + 4 \lambda \bn{\nu }_1 \\
&\leq~ 20 \lambda \bn{ \nu }_1 \,,
\end{align*}
where we used Eq. \eqref{eq:triangle-2-rev} in the first step. Dropping the first (non-negative) term on the left hand side yields
\begin{align} \label{eq:case2-rev}
\lambda \bn{ \tdelta - \hdelta }_2 ~\leq~ \lambda \bn{ \tdelta - \hdelta }_1 ~\leq~ 5 \lambda \bn{ \nu }_1 \,.
\end{align}
Combining the inequalities from Eq. (\ref{eq:case1-rev}) and (\ref{eq:case2-rev}), the following holds in both cases on $\J$,
\begin{align}
\label{eq:combine-rev}
\lambda \bn{ \tdelta - \hdelta }_2 ~\leq~ \frac{25}{16 \sqrt{s} \ngold} \bn{ \Xg \nu }_2^2 + 5\lambda \bn{ \nu }_1 + \frac{21 \lambda^2 \sqrt{s}}{\phi_R^2} \,.
\end{align}
\Halmos
\endproof

\subsection{Proof of Theorem \ref{thm:rev}}

We now use Lemma \ref{lem:rev-1} to prove Theorem \ref{thm:rev}, re-using Lemmas \ref{lem:bound-nu}, \ref{lem:J} and \ref{lem:I}.
\proof{Proof of Theorem \ref{thm:rev}}
Applying Lemma \ref{lem:bound-nu} to Lemma \ref{lem:rev-1}, the following holds with probability 1 when the events $\J$ and $\I$ hold, and $\lambda \geq 5 \lambda_0$:
\begin{align*}
\bn{ \tdelta - \hdelta }_2 ~\leq~ \frac{25 d \lambda_1}{16 \sqrt{s} \psi^2 \nproxy^2 \lambda} + \frac{5\sqrt{d\lambda_1}}{\psi \nproxy} + \frac{21 \lambda \sqrt{s}}{\phi_R^2}  \,.
\end{align*}
Then, applying Lemmas \ref{lem:J} and \ref{lem:I}, and setting $\lambda = \lambda_0/5$, we can write
\begin{align*}
\Pr\bb{\bn{ \tdelta - \hdelta }_2 ~\geq~ \frac{25 d \lambda_1}{16 \sqrt{s} \psi^2 \nproxy^2 \lambda} + \frac{5\sqrt{d\lambda_1}}{\psi \nproxy} + \frac{21 \lambda \sqrt{s}}{\phi_R^2} } &~\leq~ 1 - \Pr[\J \cap \I] \\
&~\leq~ 2d \exp\bp{-\frac{\lambda^2 \ngold}{200 \sigmag^2}} + 2d \exp\bp{-\frac{\lambda_1}{2 d \sigmap^2 \nproxy}} \,.
\end{align*}
By inspection, we again choose
\[ \lambda_1 ~=~ \frac{\nproxy^2 \lambda^2}{d} \,, \]
yielding
\begin{align*}
\Pr\bb{\bn{ \tdelta - \hdelta }_2 ~\geq~ \lambda \bp{\frac{25}{16 \sqrt{s} \psi^2 } + \frac{5}{\psi } + \frac{21 \sqrt{s}}{\phi_R^2}}} &~\leq~ 2d \exp\bp{-\frac{\lambda^2 \ngold}{200 \sigmag^2}} + 2d \exp\bp{-\frac{\lambda^2 \nproxy}{2 d^2 \sigmap^2}} \,.
\end{align*}
Finally, reversing our variable transformation as before yields the result.
\Halmos
\endproof

\section{Useful Lemmas} \label{app:lemmas}

\subsection{Properties of Gaussians}

\begin{lemma}\label{lem:l1-mu0-gaussian}
Consider a zero-mean multivariate gaussian random variable, $z \sim \N(0,\Sigma)$. Then, 
\[ \E\bb{ \bn{z}_1} ~\geq~  \sqrt{\frac{2}{\pi}} \tr\bp{\Sigma^{1/2}} \,.\]
\end{lemma}

\proof{Proof of Lemma \ref{lem:l1-mu0-gaussian}}
We begin with the case where $z$ is a scalar ($d=1$) and $\Sigma = \sigma^2$. Then, we can write
\begin{align*}
\E[|z|] &= \int_{-\infty}^{\infty} |z'| \frac{1}{\sqrt{2\pi \sigma^2}} e^{-\frac{z'^2}{2\sigma^2}} dz' \\
&= 2 \int_{0}^{\infty} |z'| \frac{1}{\sqrt{2\pi \sigma^2}} e^{-\frac{z'^2}{2\sigma^2}} dz' \\
&= \sqrt{\frac{2\sigma^2}{\pi}} \int_0^\infty e^{-u} du \\
&= \sqrt{\frac{2\sigma^2}{\pi}} \,,
\end{align*}
where we have used a variable substitution $u = \frac{z'^2}{2\sigma^2}$ (implying $du = \frac{z'}{\sigma^2} dz'$).

Then, for the case where $z$ is a vector ($d \geq 1$),
\begin{align*}
\E\bb{ \bn{z}_1} &= \sum_{i=1}^d \E[|z_i|] \\
&\geq \sum_{i=1}^d \sqrt{\frac{2\Sigma_{ii}}{\pi}}  \\
&= \sqrt{\frac{2}{\pi}} \tr\bp{\Sigma^{1/2}} \,.
\end{align*}
\Halmos
\endproof

\begin{lemma}\label{lem:l1-mu-gaussian}
Consider a multivariate gaussian random variable with mean $\mu$, $z \sim \N(\mu,\Sigma)$. Then, 
\[ \E\bb{ \bn{z}_1} ~\geq~ \max \bc{ \frac{1}{2} \bn{\mu}_1, \frac{1}{\sqrt{2\pi}} \tr\bp{\Sigma^{1/2}}} \,.\]
\end{lemma}

\proof{Proof of Lemma \ref{lem:l1-mu-gaussian}}
We begin with the case where $z$ is a scalar ($d=1$), so $\Sigma = \sigma^2$. Without loss of generality, assume $\mu \geq 0$; if not, we can equivalently consider $-z$ instead of $z$, since $|-z| = |z|$ and $-z \sim \N(-\mu, \sigma^2)$. Next, observe that $|z + \mu| \geq |z|$ if $z \geq 0$, so we can write
\begin{align*}
\E[|z|] &= \int_{-\infty}^{\infty} |z' + \mu| \frac{1}{\sqrt{2\pi \sigma^2}} e^{-\frac{z'^2}{2\sigma^2}} dz' \\
&\geq \int_{0}^{\infty} |z' + \mu| \frac{1}{\sqrt{2\pi \sigma^2}} e^{-\frac{z'^2}{2\sigma^2}} dz' \\
&\geq \int_0^\infty z' \frac{1}{\sqrt{2\pi \sigma^2}} e^{-\frac{z'^2}{2\sigma^2}} dz' \\
&= \sqrt{\frac{\sigma^2}{2\pi}} \int_0^\infty e^{-u} du \\
&= \sqrt{\frac{\sigma^2}{2\pi}} \,,
\end{align*}
where we have used a variable substitution $u = \frac{z'^2}{2\sigma^2}$ (implying $du = \frac{z'}{\sigma^2} dz'$).

In addition, observe that $|z + \mu| \geq |\mu|$ if $z \geq 0$, so we can write
\begin{align*}
\E[|z|] &= \int_{-\infty}^{\infty} |z' + \mu| \frac{1}{\sqrt{2\pi \sigma^2}} e^{-\frac{z'^2}{2\sigma^2}} dz' \\
&\geq \int_{0}^{\infty} |z' + \mu| \frac{1}{\sqrt{2\pi \sigma^2}} e^{-\frac{z'^2}{2\sigma^2}} dz' \\
&\geq |\mu| \int_0^\infty \frac{1}{\sqrt{2\pi \sigma^2}} e^{-\frac{z'^2}{2\sigma^2}} dz' \\
&= \frac{1}{2} |\mu| \,.
\end{align*}

Then, for the case where $z$ is a vector so $d \geq 1$, we can write
\begin{align*}
\E\bb{ \bn{z}_1} &= \sum_{i=1}^d \E[|z_i|] \\
&\geq \sum_{i=1}^d \max \bc{ \frac{1}{2} |\mu_i|, \sqrt{\frac{\Sigma_{ii}}{2\pi}} } \\
&\geq \max \bc{ \sum_{i=1}^d \frac{1}{2} |\mu_i|,  \sum_{i=1}^d \sqrt{\frac{\Sigma_{ii}}{2\pi}}} \\
&= \max \bc{ \frac{1}{2} \bn{\mu}_1, \frac{1}{\sqrt{2\pi}} \tr\bp{\Sigma^{1/2}}} \,.
\end{align*}
\Halmos
\endproof

\begin{lemma}\label{lem:exp-gaussian}
Consider a multivariate gaussian random variable $x\sim\mathcal{N}(\mu,\Sigma)$. Then,
\[\E\bb{\bn{z}_2^2} = \bn{\mu}_2^2+\tr\bp{\Sigma} \,.\]
\end{lemma}
\proof{Proof of Lemma \ref{lem:exp-gaussian}}
\begin{align*}
  \mathbb{E}\bb{\bn{z}_2^2} &= \E\bb{\bn{\mu+(z-\mu)}_2^2} \\
  &=\bn{\mu}_2^2+\E\bb{(z-\mu)^\top(z-\mu)} \\
  &=\bn{\mu}_2^2+\tr\bp{\E\bb{(z-\mu)(z-\mu)^\top}} \\
  &=\bn{\mu}_2^2+\tr\bp{\Sigma}.
\end{align*}
\Halmos
\endproof

\subsection{Properties of Subgaussians}

\begin{lemma}[Concentration Inequality for Subgaussians] \label{lem:subg-conc} Let $z$ be a $\sigma$-subgaussian random variable (see Definition \ref{def:subgaussian}). Then, for all $t \geq 0$,
\[ \Pr\bb{| z| ~\geq~ t} ~\leq~ 2\exp\bp{-\frac{t^2}{2\sigma^2}} \,.\]
\end{lemma}

\proof{Proof of Lemma \ref{lem:subg-conc}} See Eq. (2.9) of \cite{wainwright}.
\Halmos
\endproof

\begin{lemma}[Hoeffding Bound for Subgaussians] \label{lem:subg-hoeffding} Let $\{z_i\}_{i=1}^n$ be a set of independent $\sigma$-subgaussian random variables (see Definition \ref{def:subgaussian}). Then, for all $t \geq 0$,
\[ \Pr\bb{\sum_{i=1}^n z_i ~\geq~ t} ~\leq~ \exp\bp{-\frac{t^2}{2n\sigma^2}} \,.\]
\end{lemma}

\proof{Proof of Lemma \ref{lem:subg-hoeffding}} See Proposition 2.1 of \cite{wainwright}.
\Halmos
\endproof

\begin{lemma}\label{lem:subg-sum} Let $\{z_i\}_{i=1}^n$ be a set of independent $\sigma$-subgaussian random variables, and let $\{a_i\}_{i=1}^n$ be constants that satisfy $A = \sum_{i=1}^n a_i^2$. Then, 
\[W = \sum_{i=1}^n a_i z_i \]
is a $\bp{\sigma \sqrt{A}}$-subgaussian random variable as well.
\end{lemma}

\proof{Proof of Lemma \ref{lem:subg-sum}}
\begin{align*}
\E\bb{\exp\bp{tW}} &= \E\bb{\exp\bp{t \sum_{i=1}^n a_i z_i}} \\
&= \prod_{i=1}^n\E\bb{\exp\bp{ta_i z_i}} \\
&\leq \prod_{i=1}^n\exp\bp{\frac{\sigma^2 t^2a_i^2 }{2}} \\
&= \exp\bp{\frac{\sigma^2 t^2}{2} \sum_{i=1}^n a_i^2} \\
&= \exp\bp{\sigma^2 t^2 A /2} \,,
\end{align*}
implying that $W$ is $\bp{\sigma \sqrt{A}}$-subgaussian by Definition \ref{def:subgaussian}.
\Halmos
\endproof

\subsection{Miscellaneous} 

The following lemma is a simplification of a more general result in \cite{nesterov1998introductory}, and shows that a uniform lower bound on the second derivative of a real-valued function implies strong convexity.

\begin{lemma} \label{lem:strong-convex}
Consider a twice-differentiable function $f: \R \rightarrow \R$ with a uniform lower bound on its second derivative for all points in its domain $\mathcal{X} \subset \R$, i.e.,
\[ \inf_{x \in \mathcal{X}}f''(x) \geq m \,,\]
Then, for all $x,y \in \mathcal{X}$, $f$ satisfies
\[ f(y) - f(x) ~\geq~ f'(x)\cdot (y-x) +\frac{m}{2} \cdot (y-x)^2 \,.\] 
\end{lemma}

\proof{Proof of Lemma \ref{lem:strong-convex}}
Note that 
\[ f(y) - f(x) = \int_x^y f'(t) dt \quad\quad \text{and} \quad\quad f'(t) = f'(x) + \int_x^t f''(s) ds \,.\]
Then, we can write
\begin{align*}
f(y) - f(x) &= \int_x^y \bp{ f'(x) + \int_x^t f''(s) ds }dt \\
&= f'(x)(y-x) + \int_x^y \int_x^t f''(s) ds dt \\
&\geq f'(x) (y-x) + m \cdot \int_x^y \int_x^t ds dt \\
&= f'(x)(y-x) + \frac{m}{2} \cdot (y-x)^2 \,.
\end{align*}
\Halmos
\endproof

Following \cite{candes}, we use the following lemma to convert between $\ell_1$ and $\ell_2$ norms.
\begin{lemma} \label{lem:ell-q}
Let $b_1 \geq b_2 \geq \cdots \geq 0$. For $r \in \{1, \cdots\}$, we have
\[ \bp{\sum_{j \geq r+1} b_j^2}^{1/2} ~\leq~ r^{-1/2}\bn{b}_1 \,. \]
\end{lemma}

\proof{Proof of Lemma \ref{lem:ell-q}} Set $q=2$ in Lemma 6.9 of \cite{buhlmann}. \Halmos
\endproof

\section{Experimental Details \& Results}

\subsection{Additional Simulations} \label{app:synth-exp}

We now simulate several additional benchmarks on synthetic data.
\begin{enumerate}
\item \textit{High dimension:} The setting in Section \ref{ssec:synth} is relatively low-dimensional since $\ngold > d$. We investigate the high-dimensional setting, where $\ngold < d < \nproxy$, by taking $d=200, \ngold = 150$ and $\nproxy =  n_{test} = 2000$. All other parameters and data-generating processes remain the same as in Section \ref{ssec:synth}.
\item \textit{Theoretically-tuned regularization parameter:} Corollary \ref{cor:joint} suggests taking the regularization parameter (in the linear case) to be $\blambda$, where
\[ \blambda ~=~ C\max\bc{\sqrt{\frac{\sigmag^2 \log\bp{6bd\ngold}}{\ngold}},~ \sqrt{\frac{2d^2 \sigmap^2 \log\bp{6bd\nproxy}}{\nproxy}}}  \,.\]
We tune the multiplicative constant $C$ \textit{once} on a validation set (based on data generated as described in Section \ref{ssec:synth}) and set $C=1/10$; unlike cross-validation, this choice of regularization parameter is now held fixed across all randomly-generated datasets (i.e., iterations) and parameter values (i.e., when the values of $\ngold, \nproxy, d$ change in the high-dimensional setting). Since we are using synthetically generated data, we know the appropriate values of $\sigmag, \sigmap$ and $b$. We investigate the performance of this choice of regularization parameter compared to its cross-validated counterpart using the \textsf{glmnet} package\footnote{To match the normalization in the implementation of \textsf{glmnet}, we must divide our regularization parameter by $\ngold$.} in R.
\item \textit{One-step joint estimator:} In Section \ref{ssec:remarks}, we described a ``one-step" version of the joint estimator that simultaneously estimates the proxy and gold parameters, while enforcing an $\ell_1$ penalty on the difference of the two parameters. We implemented this estimator using the \textsf{CVXR} package in R, and tuned the regularization parameter via cross-validation on a validation set.
\end{enumerate}

\begin{figure}
\centering
  \begin{subfigure}[b]{.85 \textwidth}
    \includegraphics[width=\textwidth]{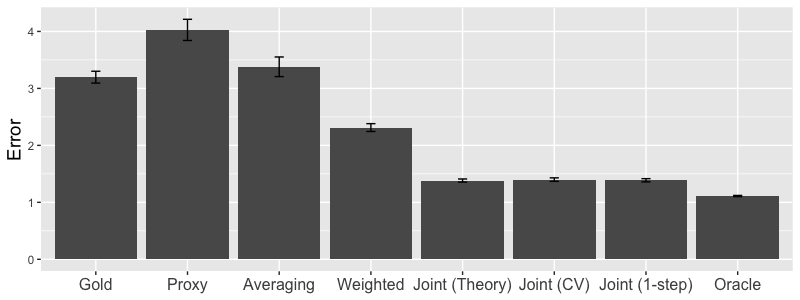}
    \caption{Low Dimension}
    \label{fig:synth-lowd}
  \end{subfigure} \\
  \begin{subfigure}[b]{.85 \textwidth}
    \includegraphics[width=\textwidth]{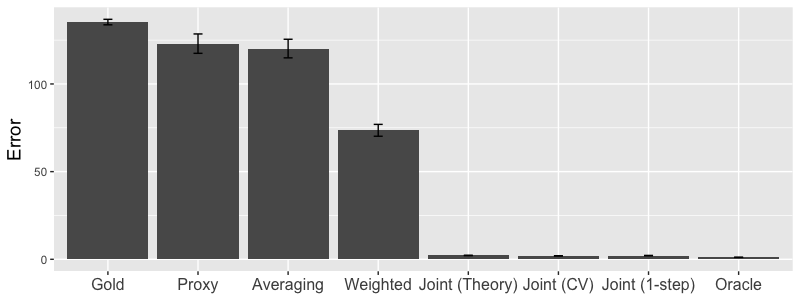}
    \caption{High Dimension}
    \label{fig:synth-larged}
  \end{subfigure}
  \caption{Out-of-sample prediction error and 95\% confidence intervals of different estimators on synthetic data.}
\end{figure}

Figure \ref{fig:synth-lowd} shows results for the low-dimensional setting (described in Section \ref{ssec:synth}) and Figure \ref{fig:synth-larged} shows results for the high-dimensional setting (described above). First, we note that the performance gain from using the joint estimator rather than the baseline estimators (gold, proxy, averaging and weighted) is far larger in high dimension compared to low dimension. This is to be expected, since it becomes more important to leverage sparse structure as $d$ grows large. Second, in both low and high dimension, we find that the performance of the joint estimator with the theoretically-tuned regularization parameter [denoted ``Joint (Theory)"] closely matches the performance of the joint estimator with the cross-validated regularization parameter [denoted ``Joint (CV)"]. This empirically validates the form of the regularization parameter proposed in Corollary \ref{cor:joint}. However, in practice, one does not know problem-specific values ($\sigmag, \sigmap,b$) needed to compute $\blambda$, making cross-validation attractive. Third, in both low and high dimension, we find little to no improvement from combining the estimation of $\hbetap$ and $\hbetag$ using the one-step joint estimator [denoted ``Joint (1-step)"]. As we argued in Section \ref{ssec:remarks}, we expect this to be the case when $\nproxy \gg \ngold$ (as we have here), since the one-step estimation decouples into our two-step joint estimator.

\subsection{Expedia Case Study Details} \label{app:expedia}

The original dataset has 9,917,530 impressions. A subset of this data includes impressions where the hotels were randomly sorted, i.e., when the provided feature random\_bool = 1. As recommended, we restrict ourselves to this subset to avoid the position bias of the existing algorithm. This results in 2,939,652 impressions.

There are 54 columns in the data. We drop the following columns:
\begin{enumerate}
\item Features that are missing more than 25\% of their entries
\item Unique identifiers for the search query, property, customer, country of property, country of customer, and search query destination
\item Time of search
\item Boolean used to identify the subset of impressions that were randomly sorted
\item Number of rooms and nights, after being used to normalize the overall price per room-night
\end{enumerate}
There are 15 remaining features and 2 outcome variables (clicks and bookings). These include: the property star rating (1-5), average customer review rating for the property (1-5), an indicator for whether the hotel is part of a major hotel chain, two different scores outlining the desirability of the hotel's location, the logarithm of the mean price of the hotel over the last trading period, the hotel position on Expedia's search results page, the displayed price in USD of the hotel for the given search, an indicator whether the hotel had a displayed sale price promotion, the length of stay, the number of days in the future the hotel stay started from the search date, the number of adults, the number of children, the number of rooms, and an indicator for a weekend stay.
We also drop impressions that have missing values and outliers at the 99.99\% level, leaving 2,262,166 total impressions. As recommended by \cite{friedman2001}, we standardize each feature before performing any regressions.

As described in Section \ref{ssec:expedia}, we use a subsample of 10,000 randomly drawn observations in each iteration. We first reserve 50\% of this data as a held-out test set to assess performance. The remaining 50\% is used as the training set for the gold and proxy estimators. For the averaging, weighted, and joint estimators, we additionally need to choose a tuning parameter. Following standard practice, we use a random 70\% subsample of the observations (that are not in the test set) as our training set and the remaining 30\% as a validation set. We train models with different values of $\lambda$ on the training set, and use mean squared prediction error on the validation set to choose the best value of $\lambda$ in the final model. Finally, the ``Oracle" is trained on all 2M+ impressions excluding the test set.

\subsection{Diabetes Case Study Details} \label{app:emr}

The original dataset has 9948 patient records across 379 healthcare providers. The data only contains patients who have recently visited the provider at least twice. Each patient is associated with 184 features constructed from patient-specific information available \textit{before} the most recent visit (i.e., indicator variables for past ICD-9 diagnoses, medication prescriptions, and procedures), as well as a binary outcome variable \textit{from} the most recent visit (i.e., whether s/he was diagnosed with diabetes in the last visit). As described in Section \ref{ssec:diabetes}, we only study 3 of the 379 providers: we use patient data from a medium-sized provider as our gold data, and patient data pooled from two larger providers as our proxy data.

However, we use patient data from the remaining 376 providers to do variable selection as a pre-processing step. In particular, we run a simple LASSO variable selection procedure by regressing diabetes outcomes against the 184 total features (note that we exclude the 3 healthcare providers that constitute the proxy and gold populations in this step to avoid overfitting). This leaves us with roughly 100 commonly predictive features (depending on the randomness in the cross-validation procedure).

We first reserve 50\% of the gold data as a held-out test set to assess performance, and use the remaining 50\% of the gold data for training models. Unlike the Expedia case study, our test set does not overlap with the proxy data (since the gold and proxy data are derived from different cohorts) so the entire proxy data is used for training. The gold estimator is trained on all gold observations that are not in the test set; the proxy estimator is trained on all proxy observations. For the averaging, weighted, and joint estimators, we additionally need to choose a tuning parameter. Following standard practice (modified to accommodate proxy data), we use a random 70\% subsample of the gold observations (that are not in the test set) and all of the proxy observations as our training set; we use the remaining 30\% of the gold observations as our validation set. We train models with different values of $\lambda$ on the training set, and use mean squared prediction error on the validation set to choose the best value of $\lambda$ in the final model.

\end{APPENDICES}

\end{document}